\documentclass{article}
\usepackage{natbib}

\usepackage[utf8]{inputenc} %
\usepackage[T1]{fontenc}    %
\usepackage{url}            %
\usepackage{booktabs}       %
\usepackage[hidelinks]{hyperref}       %
\usepackage{amsfonts}       %
\usepackage{nicefrac}       %
\usepackage{microtype}      %
\usepackage{xcolor}   
\usepackage{bbm}
\usepackage{wrapfig}

\usepackage{placeins}

\usepackage[final]{neurips_2022}

\usepackage[pdftex]{graphicx}
\graphicspath{{images/}}

\usepackage{parskip}

\usepackage{subcaption}
\usepackage{csquotes}
\usepackage{enumitem}

\usepackage[smallcaps, acronym]{glossaries}
\newacronym{sbi}{sbi}{simulation-based inference}
\newacronym{abc}{abc}{approximate Bayesian computation}
\newacronym{npe}{npe}{neural posterior estimation}
\newacronym{nnpe}{nnpe}{noisy neural posterior estimation}
\newacronym{nle}{nle}{neural likelihood estimation}
\newacronym{rnpe}{rnpe}{robust neural posterior estimation}
\newacronym{mcmc}{mcmc}{Markov chain Monte Carlo}
\newacronym{mmd}{mmd}{maximum mean discrepancy}
\newacronym{hmc}{hmc}{Hamiltonian Monte Carlo}

\newcommand{\pindent}{\parshape 2 0pt \linewidth 0.6cm \dimexpr\linewidth-0.6cm}

\title{Robust Neural Posterior Estimation and \\ Statistical Model Criticism}

\usepackage{authblk}

\title{\textbf{Robust Neural Posterior Estimation and Statistical Model Criticism}}
\author[1]{Daniel Ward}
\author[2]{Patrick Cannon}
\author[3]{Mark Beaumont}
\author[1]{Matteo Fasiolo}
\author[2,4]{Sebastian~M.~Schmon}
\affil[1]{School of Mathematics, Bristol University, UK}
\affil[2]{Improbable, UK}
\affil[3]{School of Biological Sciences, Bristol University, UK}
\affil[4]{Department of Mathematical Sciences, Durham University, UK}
\date{}                     %
\usepackage[ruled]{algorithm2e} %
\usepackage{xcolor} %
\usepackage{tikz}
\usetikzlibrary{fit,calc}

\makeatletter
\renewcommand{\algocf@makecaption}[2]{%
  \parbox[t]{\columnwidth}{\algocf@captiontext{#1}{#2}}%
}%
\renewcommand{\algocf@makecaption@boxed}[2]{%
  \global\sbox\algocf@capbox{\algocf@makecaption{#1}{#2}}%
 }%
\renewcommand{\algocf@caption@boxed}{\vskip\AlCapSkip
  \leavevmode\hskip-\leftskip\box\algocf@capbox\hskip-\rightskip}
\renewcommand{\@algocf@capt@plain}{above}%
\makeatother

\usepackage{amsmath}
\newcommand{\E}{\mathbb{E}}
\usepackage{bm}
\newcommand{\+}[1]{\bm{#1}}  %

\begin{document}

\maketitle

\begin{abstract}
Computer simulations have proven a valuable tool for understanding complex phenomena across the sciences. However, the utility of simulators for modelling and forecasting purposes is often restricted by low data quality, as well as practical limits to model fidelity. 
In order to circumvent these difficulties, we argue that modellers must treat simulators as idealistic representations of the true data generating process, and consequently should thoughtfully consider the risk of \emph{model misspecification}. 
In this work we revisit \gls{npe}, a class of algorithms that enable black-box parameter inference in simulation models, and consider the implication of a simulation-to-reality gap. 
While recent works have demonstrated reliable performance of these methods, the analyses have been performed using synthetic data generated by the simulator model itself, and have therefore only addressed the well-specified case. In this paper, we find that the presence of misspecification, in contrast, leads to unreliable inference when \gls{npe} is used na\"ively. As a remedy we argue that principled scientific inquiry with simulators should incorporate a \emph{model criticism} component, to facilitate interpretable identification of misspecification and a \emph{robust inference} component, to fit \enquote*{wrong but useful} models. We propose \gls{rnpe}, an extension of \gls{npe} to simultaneously achieve both these aims, through explicitly modelling the discrepancies between simulations and the observed data. We assess the approach on a range of artificially misspecified examples, and find \gls{rnpe} performs well across the tasks, whereas na\"ively using \gls{npe} leads to misleading and erratic posteriors. \\

\end{abstract}
\glsreset{npe}
\glsreset{rnpe}
\glsreset{sbi}

\section{Introduction}

Stochastic simulators have become a ubiquitous modelling tool across the sciences and are regularly applied to some of the most complex and challenging problems of scientific interest, including climate change \citep[see e.g.][]{randall2007climate}, particle physics \cite[e.g.][]{brehmer2020madminer}, and the Covid-19 pandemic \citep[e.g.][]{ferguson2020impact}. Simulators implicitly define a likelihood function $p(\+x \mid \+\theta)$, where $\+x$ is the simulator output and $\+\theta$ are the simulator parameters. Although running the simulator to sample from the model is straightforward, the inherent complexity of simulators often makes analytic calculation of the likelihood intractable. As a result, classical inference techniques to find the parameter posterior $p(\+ \theta \mid \+x)$ such as \gls{mcmc} \citep{metropolis1953equation} are infeasible. To overcome this issue, a large family of \gls{sbi} methods have been developed that allow parameter inference to be performed on arbitrary black-box simulators \citep[see][]{cranmer2020frontier}. Broadly, these approaches estimate a function that allows access to an approximate posterior. This can be achieved by approximating the posterior directly \citep{papamakarios2016fast, greenberg2019automatic, lueckmann2017flexible}, or indirectly, via approximating the likelihood \citep{papamakarios2019sequential} or likelihood-to-evidence ratio \citep{hermans2020likelihood, thomas2022likelihood}, from which the posterior can be sampled using \gls{mcmc}. For estimating the posterior or likelihood, \gls{npe} and \gls{nle} have shown to be powerful approaches \citep{lueckmann2021benchmarking}, which rely on neural network-based conditional density estimators, such as normalising flows, to approximate the likelihood or posterior \citep{papamakarios2019normalizing}.

\begin{figure}[h]
    \centering
    \includegraphics{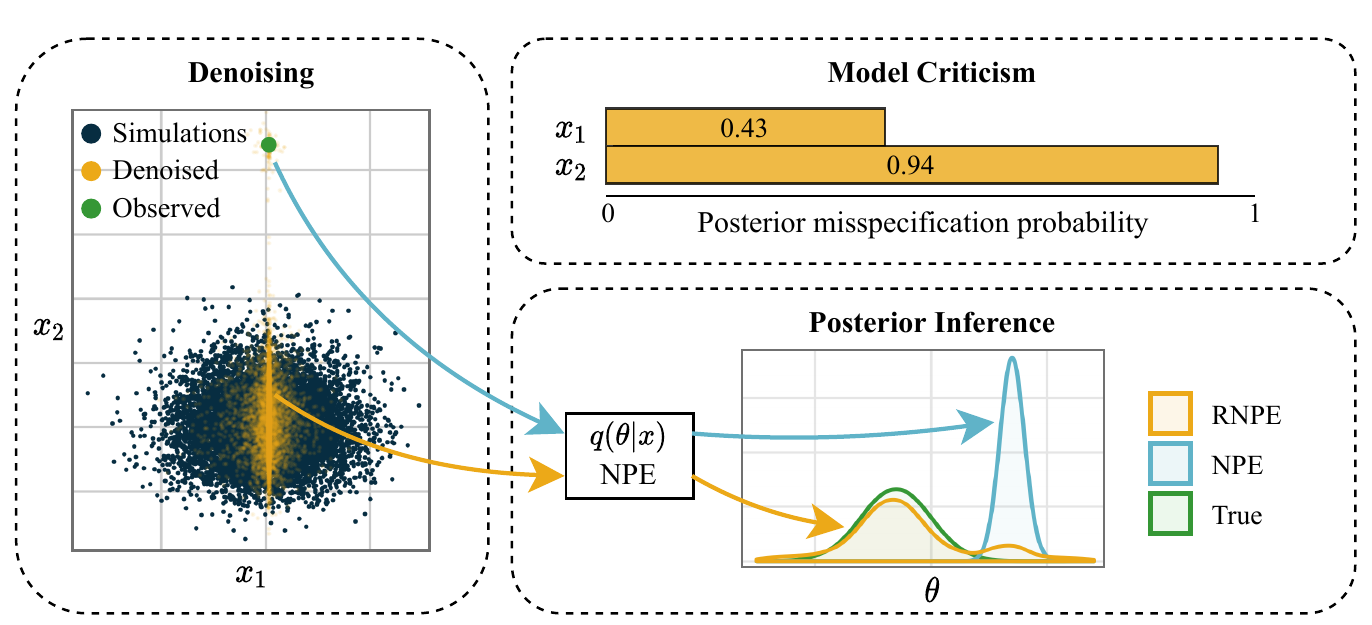}
    \caption{Overview of the \gls{rnpe} framework. Through \enquote*{denoising} the observed data, we can simultaneously perform \emph{model criticism} to identify how the model is misspecified, and perform inference \emph{robust} to model misspecification. See Gaussian example in section \ref{section:tasks} for details of the experiment.}
    \label{fig:diagram}
\end{figure}
\glsreset{rnpe} %

To define misspecification, it is necessary to make a clear distinction between the true data generating process and the simulator: we will write $\+y_o$ for observed data, following some unknown data generating process with distribution denoted $p^*$, and we will write $\+x \sim p(\+x \mid \+\theta)$ to denote samples from the simulator. 
In many applications of \gls{sbi}, summary statistics are used to reduce the dimensionality of the problem. If this is the case, we use $\+x$ and $\+y_o$ to denote the simulated and observed summary statistics, rather than the raw data, and treat the raw data as an unobserved latent variable of the model.
The simulator is said to be \emph{misspecified} if the true data generating process does not fall within the family of distributions defined by the simulator, i.e. $p^* \notin \{ p(\+x \mid \+\theta);\ \+\theta \in \+\Theta \}$. Simulators generally form idealised and simplistic representations of a more complex data generating process, and hence~\textendash{}~like all statistical models{  }\textendash{  }will be wrong to some extent. This misspecification often results in a discrepancy between simulated and observed data, sometimes termed a \enquote{simulation-to-reality gap} \citep[e.g.][]{miglino1995evolving}. Despite this, the general approach in \gls{sbi} is to learn a posterior approximation $q(\+\theta \mid \+x)$ (either directly, or indirectly), using simulated data, $(\+x, \+\theta) \sim p(\+x \mid \+\theta)p(\+\theta)$, and to condition on the observed data, $q(\+\theta \mid \+x=\+y_o)$. This approach is only theoretically justified under the assumption that the simulator is well-specified; relying on this approach for misspecified simulators leads to two issues:

\begin{enumerate}[label=(\roman*)]
    \item As the posterior approximation $q(\+\theta \mid \+x)$ is learned using samples from the simulator, we can only expect it to form a reasonable approximation of $p(\+\theta \mid \+x)$ for regions well covered by simulations. For misspecified simulators, the observed data $\+y_o$ may be very unlikely in $p(\+x) = \int p(\+x|\+\theta)p(\+\theta) d\+\theta$, in which case the approximation $q(\+\theta \mid \+x = \+y_o)$ often becomes poor \citep{withcannonyoucan}. In some cases, $\+y_o$ may even fall outside the support of $p(\+x)$,  which results in \gls{npe} attempting to extrapolate to estimate the undefined posterior.
    \item It has been demonstrated more generally that Bayesian inference is frequently problematic under misspecification. Even for simple models, misspecification can lead to posterior concentration around \enquote{bad} models \citep{grunwald2007suboptimal, grunwald2017inconsistency} and poorly calibrated credible regions \citep{kleijn2012bernstein, syring2019calibrating}. These issues will persist even in the unlikely case that $q(\+\theta \mid \+x)$ perfectly approximates $p(\+\theta \mid \+x)$.
\end{enumerate}

One approach to improve robustness is to incorporate an error model to explicitly model potential discrepancies between the observed data and simulations. A simple method would be to directly incorporate the error model within the simulator itself, such as by applying additive noise to the output to ensure that the observed data has reasonable marginal likelihood under the model \citep{cranmer2020frontier}. However, this approach has some limitations: \emph{i)} it is challenging to interpret how the model is misspecified -- an obvious choice is to use posterior predictive checks, but poor predictive performance could result from failed inference, limitations of simulator, or both; \emph{ii)} retraining of the approximate posterior would be required to investigate different error models, which can be computationally expensive; and \emph{iii)} the error model itself is likely tractable, which can be utilised to improve inference. 
\paragraph*{Our contribution.}
We propose \gls{rnpe}, an \gls{sbi} framework that incorporates an error model, $p(\+y\mid\+x)$ to account for discrepancies between simulations and the observed data, whilst avoiding the limitations of na\"ively incorporating it directly into the simulator. We explicitly learn to invert the error model for the observed data prior to inference of the parameter posterior, a process we call \enquote*{denoising}. This approach yields the following advantages:

\pindent
\textbf{Efficient Model Criticism.} Understanding a model's limitations is crucial for principled model development \citep{box1980sampling, gelman2013philosophy}. By probabilistically inverting the error process for the observed data, and inferring any associated error model parameters, we can detect and interpret discrepancies between the simulated and observed data, allowing criticism of the simulator. This process can be achieved in a manner decoupled from inference, which provides efficiency benefits, whilst avoiding confounding inference failures with problems arising due to limitations of the simulator.

\pindent
\textbf{Robust Inference.} Even after several rounds of model development and model criticism, the best available model will still frequently be misspecified to an appreciable degree. However, it is often still desirable to use \enquote*{wrong but useful} models for parameter inference and predictions, and hence the inference process should be robust to misspecification.
By exploiting the amortisation\footnote{Amortisation refers to the ability of a conditional distribution approximation to condition on arbitrary instances of the conditioning variable, rather than being specialised to a particular instance.} property of \gls{npe}, we can ensemble posterior estimates over a set of \enquote*{denoised} observations, leading to posterior estimates robust to misspecification.

An overview of the \gls{rnpe} framework can be seen in Fig. \ref{fig:diagram}. We examine the practical value of this approach for \emph{model criticism} and \emph{robust inference} on examples in which we artificially introduce realistic levels of misspecification.

\section{Method}

\subsection{Error Model}
\label{section:error_model}
As outlined above, we can explicitly model discrepancies between the observed data and simulations by introducing an error model $p(\+y\mid\+x)$, such that the assumed generative model is
\begin{align}
    \label{eq:gen_model}
    p(\+\theta, \+x, \+y) &= p(\+y \mid \+x)p(\+x \mid \+\theta)p(\+\theta),
\end{align}
where we assume that the error is independent of the simulator parameters $\+\theta$, and we treat $\+x$ as an unobserved latent variable of the model. Using the equivalent factorisation
$$p(\+\theta, \+x, \+y)=p(\+\theta \mid  \+x)p(\+x \mid \+y)p(\+y),$$
we can marginalise over the latent $\+x$ and find an expression for the parameter posterior 
\begin{align}
    p(\+\theta, \+y)&=p(\+y) \int p(\+x \mid \+y)p(\+\theta \mid  \+x) \, \mathrm{d} \+x, \\
    p(\+\theta \mid \+y) &= \int p(\+x \mid \+y)p(\+\theta \mid  \+x) \, \mathrm{d} \+x = \E_{\+x \sim p(\+x \mid \+y)}[ p(\+\theta \mid \+x)]. \label{eq:expectation}
\end{align}

Equation \eqref{eq:expectation}, implies that if we had access to $p(\+x \mid \+y)$ and $p(\+\theta \mid \+x)$, we could sample from the parameter posterior by first sampling from the posterior over the latent variables $\+x \sim p(\+x \mid \+y)$, and then sampling from $p(\+\theta \mid \+x)$. This approach reverses the data generating process from Equation \eqref{eq:gen_model}, whilst propagating uncertainty throughout, first removing the error from the observed data by denoising, and then finding the associated simulator parameters given the denoised data. A Monte Carlo approximation of the expectation in Equation \eqref{eq:expectation} can also be used to estimate the posterior density $p(\+\theta \mid \+y)$, i.e.
\begin{align}
    p(\+\theta \mid \+y) \approx \frac{1}{M}\sum_{m=1}^M p(\+\theta \mid \tilde{\+x}_m), \quad \quad \tilde{\+x}_1, \ldots , \tilde{\+x}_M, \overset{\mathrm{iid}}{\sim} p(\+x \mid \+y),
\end{align}
however, we note that in some scenarios this approximation could have high variance. In practice, $p(\+x \mid \+y)$ and $p(\+\theta \mid \+x)$ are unknown. We can approximate $p(\+\theta\mid\+x)$ by training a normalising flow $q(\+\theta \mid \+x)$ on simulated data $(\+x, \+\theta) \sim p(\+x \mid \+\theta)p(\+\theta)$, as is commonly done for \gls{npe}. To approximately sample from $p(\+x \mid \+y)$, we \emph{i)} specify an error model $p(\+y \mid \+x)$, \emph{ii)} train a normalising flow, $q(\+x)$, fitted to samples from the prior predictive distribution of the simulator $\+x \sim p(\+x)$, and \emph{iii)} sample $\tilde{\+x} \sim p(\+x \mid \+y)$, along with any error model parameters, using \gls{mcmc}, as $\hat{p}(\+x \mid \+y) \propto p(\+y \mid \+x)  q(\+x)$. We denote samples from $\hat{p}(\+x \mid \+y)$ and $q(\+\theta \mid \+x)$ as $\tilde{\+x}$ and $\tilde{\+\theta}$ to avoid confusion with prior samples and simulations. Pseudo-code for the overall approach is given in Algorithm \ref{alg:robust_npe}.

\begin{algorithm}
    \caption{Robust neural posterior estimation (\gls{rnpe})}
    \label{alg:robust_npe}
    \For{$i$ in $1:N$}{
        \nl Sample $\+\theta_i \sim p(\+\theta)$\\
        \nl Simulate $\+x_i \sim p(\+x \mid \+\theta)$ \\
    }
    \nl Train \gls{npe} $q(\+\theta \mid \+x)$ on $\{(\+\theta_i, \+x_i)\}_{i=1}^N$ \label{train_posterior} \\
    \nl Train $q(\+x)$ on $\{\+x_i\}_{i=1}^N$ \label{train_marginal_likelihood}\\
    \nl Sample $\tilde{\+x}_m \sim  \hat{p}(\+x \mid  \+y_o) \propto p(\+y_o \mid \+x) q(\+x), \  m=1, \ldots ,M$ using \gls{mcmc} \label{samp_denoise} \\
    \nl Sample $\tilde{\+\theta}_m \sim q(\+\theta \mid \tilde{\+x}_m),\ m=1, \ldots ,M$\label{samp_posterior} \\
    \textbf{return} $\{(\tilde{\+\theta}_m, \tilde{\+x}_m)\}_{m=1}^M$, samples drawn approximately from $p(\+\theta, \+x \mid \+y_o)$
\end{algorithm}

Under this framework, the standard \gls{sbi} approach can be retrieved as the special case in which $p(\+y \mid \+x)=p(\+x \mid \+y)= \delta(\+x-\+y)$, where $\delta$ is the Dirac delta distribution, meaning that we revert to assuming no discrepancy between the simulator and data generating process.

The idealised nature of simulation models implies that frequently there will be some characteristics of the observed data that are well captured by simulations, and some aspects in which there exists a discrepancy. Using a set of summary statistics for inference, rather than the raw data naturally captures an interpretable, low dimensional and diverse set of characteristics of the simulated and observed data. Because of this, we find that in general, reducing the raw data to summary statistics is advantageous for \emph{model criticism}, as the practitioner can choose summary statistics which align with their belief about important model characteristics, and easily interpret any discrepancies (see Section \ref{section:results}). The requirement for domain specific knowledge to develop summary statistics has been alleviated by recent work \citep[e.g.][]{fearnhead2012constructing, chan2018likelihood, chen2020neural}, which seek to automatically embed the raw data to summary statistics; however, these embeddings often lack a tangible interpretation. Additionally, the embedding methods themselves may not be robust to misspecification. One approach is to use both hand-crafted and automatic summary statistics in tandem. Hereafter, however, we generally perform \emph{model criticism} and \emph{inference} using hand-crafted summary statistics, such that $\+x$ and $\+y_o$ refer to the simulated and observed summary statistics, \emph{not} the raw data.

\subsection{Spike and Slab}
\label{error_model_choice}
In practice, the true error model is unknown, and hence the error model should describe our prior belief over the discrepancy between the data generating process and simulations. Additionally, to facilitate model criticism, it should ideally allow easy assessment of which summary statistics are approximately well-specified and which are misspecified. Inspired by sparsity-inducing priors used in Bayesian regression literature \citep{george1993variable}, we use a \enquote{spike and slab} error model on each summary statistic
\begin{align}
    \+x &\sim q(\+x), & \hspace{-10cm}&\\
    z_j &\sim \text{Bernoulli}(\rho), \hspace{1cm} &j=1, \ldots , D \\
    y_j  \mid x_j,  z_j &\sim 
    \begin{cases}
    \ N(x_j, \sigma^2), & \text{ if }  z_j=0\\
    \ \text{Cauchy}(x_j, \tau), & \text{ if } z_j=1
    \end{cases} \hspace{-10cm} &j=1, \ldots , D
\end{align}
where $\+x \in \mathbb{R}^D$. As the summary statistics have varying scales, we standardise them to have mean zero and variance one. The Bernoulli variables indicate whether the $j$-th summary statistic is approximately well-specified; we use the probability of $\rho = 0.5$ to express the belief that it is equally likely a summary statistic will be misspecified or well-specified \textit{a priori}. When a summary statistic is well-specified (i.e., $z_j = 0$), we take the error distribution to be a tight Gaussian (a spike) centred on $x_j$ , choosing $\sigma=0.01$ to enforce consistency between the simulator and the observed data for the $j$-th statistic. In contrast, in the misspecified case (i.e., $z_j = 1$), we take the error model to be the much wider and heavy tailed Cauchy distribution (a slab). Generally, we might expect misspecification to be relatively subtle, but some model inadequacies can lead to catastrophically misspecified summary statistics. We chose the Cauchy scale $\tau=0.25$, to reflect this, as it places half the mass within $\pm 0.25$ standard deviations of $x_j$, but the long tails accommodate summary statistics that are highly misspecified. The sparsity-inducing effect of this error model matches what we consider to be a common scenario, namely that a proportion of the summary statistics are jointly consistent with the simulator, whereas others may be incompatible. If we expect \textit{a priori} more or fewer of the summary statistics to be misspecified, the prior misspecification probability can be adjusted accordingly. By marginalising over $\+z$, the spike and slab error model can also be written as
\begin{align}
    p(\+y \mid \+x) = \prod_{j=1}^D \big[(1 - \rho) \cdot p(y_j \mid x_j, z_j=0) + \rho \cdot p(y_j \mid x_j, z_j=1)\big],
\end{align}
and as such can be viewed as an equally weighted mixture of the Gaussian spike, $p(y_j \mid x_j, z_j=0)$, and the Cauchy slab, $p(y_j \mid x_j, z_j=1)$, for each summary statistic. We note that error model hyperparameter tuning approaches could be considered, e.g. a reasonable heuristic would be to choose an error model that is broad enough that the denoised data $\tilde{\+x}$ tend not to be outliers in $q(\+x)$, compared to simulations\footnote{This could for example be assessed by estimating highest density regions of $q(\+x)$, using the density quantile approach from  \cite{hyndman1996computing}.}. However, we chose to keep the hyperparameters consistent across tasks, to avoid the risk of overfitting to the tasks and to demonstrate that neither strong prior knowledge on the error model, nor careful hyperparameter tuning is necessary to yield substantial improvements in performance.

A key advantage of the spike and slab error model is given by the latent variable $\+z$. Similar to posterior inclusion probabilities in Bayesian regression, the posterior frequency of being in the slab, $\text{Pr}(z_j = 1 \mid \+y)$, can be used as an indicator of the \emph{posterior misspecification probability} for the $j$-th summary statistic. By comparing to the prior misspecification probability $\rho$, we can say that if $\text{Pr}(z_j = 1 \mid \+y)>\rho$, the model provides evidence of misspecification for the $j$-th summary statistic, and if $\text{Pr}(z_j = 1 \mid \+y)<\rho$, it provides evidence it is well-specified  \citep{sep-epistemology-bayesian}.
For the purpose of generality, the latent variable $\+z$ was not included when describing \gls{rnpe} thus far. However, we can jointly infer the posterior $\hat{p}(\+x, \+z \mid \+y)$ in step \ref{samp_denoise} of algorithm \ref{alg:robust_npe}, by using an \gls{mcmc} algorithm that supports sampling both continuous and discrete variables. We used mixed \gls{hmc} \citep{zhou2020mixed, zhou2022metropolis} from the NumPyro python package \citep{phan2019composable}, which is an adaptation of \gls{hmc} \citep{neal2011mcmc, duane1987hybrid}, for this purpose.

\section{Related Work}
\subsection[Model Criticism in sbi]{Model Criticism in \gls{sbi}}
Posterior predictive checks have been widely used in \gls{sbi}, to compare the predictive samples to the observed data \cite[e.g.][]{durkan2020contrastive, greenberg2019automatic, papamakarios2019sequential}. Although this is a form of model criticism, we note a key limitation of this approach is that if a discrepancy is discovered, it may not be clear whether this is due to the failure of the inference procedure, or due to simulator misspecification. In general, it may be possible to identify the presence of misspecification using anomaly/novelty detection. One such approach was suggested by \cite{schmitt2021bayesflow} in the context of \gls{sbi}. However, this approach does not facilitate interpreting misspecification in the sense presented here, and it is not clear how to extend it to perform robust inference.

There are a few related methods that criticise a posterior estimate by assessing its calibration using simulated data  \citep[e.g.][]{hermans2021averting, prangle2014diagnostic, talts2018validating}; however, these approaches criticise the inference procedure, \emph{not} the simulator itself, and the associated results rely on the simulator being well-specified.

\subsection[Robust Inference in sbi]{Robust Inference in \gls{sbi}}
\Gls{abc} is a family of \gls{sbi} methods characterised by their use of rejection sampling alongside a kernel function to compare the simulated and observed data \citep{tavare1997inferring, pritchard1999population, beaumont2002approximate}. The simplest form of \gls{abc} samples candidate parameters from the prior distribution and accepts them with probability proportional to the similarity of the corresponding simulations to the observed data. The use of a kernel is often seen as a necessary evil, as enforcing an exact match between the observed and simulated data generally results in all samples being rejected. However, it has been noted that the kernel function can often be interpreted as specifying an error model distribution $p(\+y \mid \+x)$ \citep{wilkinson2013approximate}. This idea has been expanded using generalised Bayesian inference, in which the kernel is replaced with a loss function, which can be chosen to use a robust measure of discrepancy such as the \gls{mmd} \citep{cherief2020mmd, frazier2020robust, fujisawa2021gamma, park2015k2, schmon2020generalized, dyer2021approximate}.

Synthetic likelihood is an \gls{sbi} method that uses a Gaussian approximation of the likelihood \citep{wood2010statistical}. \cite{frazier2021robust} investigated a robust synthetic likelihood algorithm, in which the variance components of the Gaussian likelihood are expanded based on additional free parameters, inferred in a nested \gls{mcmc} sampling scheme. In the context of the paper, they interpret this approach as \enquote{likelihood adjustment}; however, this could equivalently be reframed as the addition of an independent Gaussian error model, with a prior on the variances\footnote{This results due the fact that the addition of two independent Gaussian variables yields another Gaussian random variable, with covariance equal to the sum of the individual covariance matrices.}.

In contrast to more traditional \gls{sbi} methods, few solutions have been proposed for performing robust inference with neural \gls{sbi} methods. Perhaps the most common approach is to augment the model with additional nuisance parameters to expand the range of values the simulator can reasonably produce \citep{cranmer2020frontier}. This again can often be interpreted as a method of incorporating an error model within the simulator. However, as previously discussed, the primary limitation of this approach is that it does not easily facilitate \emph{model criticism}.

\section{Experimental setup}
On three misspecified tasks, we benchmark the performance of three methods for parameter inference: \emph{i)} \gls{rnpe}, \emph{ii)} \gls{nnpe} in which we directly approximate $p(\+\theta \mid \+y)$ by incorporating the error model into the simulator, and \emph{iii)} \gls{npe}, in which we assume no model error is present. To reliably assess performance, we repeat the inference procedure with $1000$ different observations and ground truth parameter pairs for each misspecified task, and calculate two metrics: \emph{i)} the log probability of the true parameter vector, $\+\theta^*$, and \emph{ii)} the posterior coverage properties (see Section \ref{section:metrics}). Further metrics, and results for the well-specified case, can be found in Appendix \ref{appendix:additional_results}. We further show examples of \gls{rnpe} on a single observation for each task to demonstrate how \gls{rnpe} can be applied in practice for model criticism and robust posterior inference.

For all experiments, we used $N \hspace{-0.05cm} = \hspace{-0.025cm} 50,000$ simulations, with $M \hspace{-0.05cm} = \hspace{-0.025cm} 100,000$ \gls{mcmc} samples following $20,000$ warm up steps. The \gls{mcmc} chains were initialised using a random simulation, and $z_j=1 \ \text{for}\  j=1,\ldots , D$. To build the approximation $q(\+x)$, we used block neural autoregressive flows \citep{de2020block}, as we do not require the ability to sample directly from $q(\+x)$. For the approximation of $q(\+\theta \mid \+x)$, we used neural spline flows \citep{durkan2019neural}. For all tasks the hyperparameters were kept consistent; information on hyperparameter choices can be found in Appendix \ref{appendix:hyperparameters}. The code required to reproduce all the results from this manuscript is available at \url{https://github.com/danielward27/rnpe}.

\subsection{Tasks}
\label{section:tasks}
Detailed descriptions of each task can be found in Appendix \ref{appendix:tasks}.

\textbf{Gaussian.} A simple Gaussian example from \cite{frazier2020model}. The task involves predicting a single parameter $\mu$, using the sample mean and variance of 100 i.i.d. samples drawn from $N(\mu, 1)$, where $1$ is the variance of the Gaussian distribution. To artificially produce misspecified observations, the 100 i.i.d. are instead drawn from $N(\mu, 2)$. Despite its simplicity, this task usefully demonstrates that issues with misspecification can arise even in the simplest of models.

\textbf{Gaussian Linear.} A 10-d linear Gaussian model from \cite{lueckmann2021benchmarking}. The parameter $\+\theta$ is the mean vector of a Gaussian likelihood. Additive Gaussian noise is used to introduce misspecification. For this task, the spike and slab error model employed will be reasonably misspecified, as the noise is not sparse or heavy tailed.

\textbf{SIR.} A stochastic model of epidemic spread with a time-varying infection rate. We attempt to infer two parameters, the infection rate $\beta$ and the recovery rate $\gamma$. We computed the following summary statistics: 1-3) \texttt{Mean}, \texttt{Median} and \texttt{Max} - the mean, median and maximum number of infections; 4) \texttt{Max Day} - the day of occurrence of the maximum number of infections, 5) \texttt{Half Day} - the day at which half the total number of infections was reached, and 6) \texttt{Autocor} - the mean autocorrelation (lag 1) of infections. To artificially produce misspecified observations, we chose to introduce a small degree of reporting delays, whereby weekend infection counts are reduced by 5\%, which are subsequently recouped on the following Monday. Fig. \ref{fig:sir_example}d shows an example of a raw observation.

\textbf{CS.} A marked point process model of cancer and stromal (CS) cell development in 2D-space, in which the locality of a cell relative to unobserved parent points determines whether it is a cancer or a stromal cell. Similar point process models can be found in \cite{jones2019identifying}. There are three Poisson rate parameters, corresponding to the cell rate $\lambda_c$, the parent rate $\lambda_p$ and the daughter rate $\lambda_d$, where $\lambda_d$ controls the number of nearest cells to each parent which become cancerous. The following summary statistics were used: 1-2) \texttt{N Cancer} and \texttt{N Stromal} - the number of cancer and stromal cells; 3-4) \texttt{Mean Min Dist} and \texttt{Max Min Dist} - the mean and maximum distance  from stromal cells to their nearest cancer cell. To artificially produce misspecified observations, cancer cells were removed if they fell too close to a parent point, mimicking necrosis that often occurs in core regions of tumours. A plot of an example misspecified simulation and a corresponding artificially produced observation can be found in Fig. \ref{fig:misspecification_cancer} in the appendix.

\subsection{Performance Metrics}
\label{section:metrics}
\textbf{Log probability of $\+\theta^*$.} The mean posterior log probability of the true parameters over multiple observations is an extensively used performance metric \cite[e.g.][]{papamakarios2016fast, hermans2020likelihood, durkan2020contrastive}, and is closely related to the average KL divergence between the true and approximate posteriors \citep{lueckmann2021benchmarking}. We found that \gls{npe} would occasionally fail catastrophically, placing negligible posterior mass on the true parameters, skewing the mean estimates. Due to this, we instead use box plots to visualise the distribution over the log-probabilities of the true parameters, rather than reporting the mean.

\textbf{Posterior Coverage.} Given a confidence level $\alpha$ and a posterior approximation $\hat{p}(\+\theta \mid \+y)$, let $\operatorname{HDR}_{\hat{p}(\+\theta \mid \+y)}(1-\alpha)$ represent its $1-\alpha$ highest posterior density region, i.e. the smallest region that contains at least $100(1-\alpha)\%$ of the mass of $\hat{p}(\+\theta \mid \+y)$. The expected posterior coverage is the frequency with which the true parameter value falls within this highest density region
\begin{align}
    \E_{\+\theta^*, \+y \sim p(\+\theta, \+y)}\left[\mathbbm{1}\{\+\theta^* \in \operatorname{HDR}_{\hat{p}(\+\theta \mid \+y)}(1-\alpha)\}\right]
\end{align}
where $\mathbbm{1}$ is the indicator function. Posterior coverage is useful for assessing whether an inference procedure is likely to yield overconfident or conservative posterior estimates \citep{hermans2021averting}. We empirically estimate this expectation, approximating the highest density regions using the density quantile approach outlined in \cite{hyndman1996computing}, with 10,000 samples from the approximate posterior.

\section{Results and Discussion}
\label{section:results}

\begin{figure}[t]
    \centering
    \includegraphics[width=0.85\linewidth]{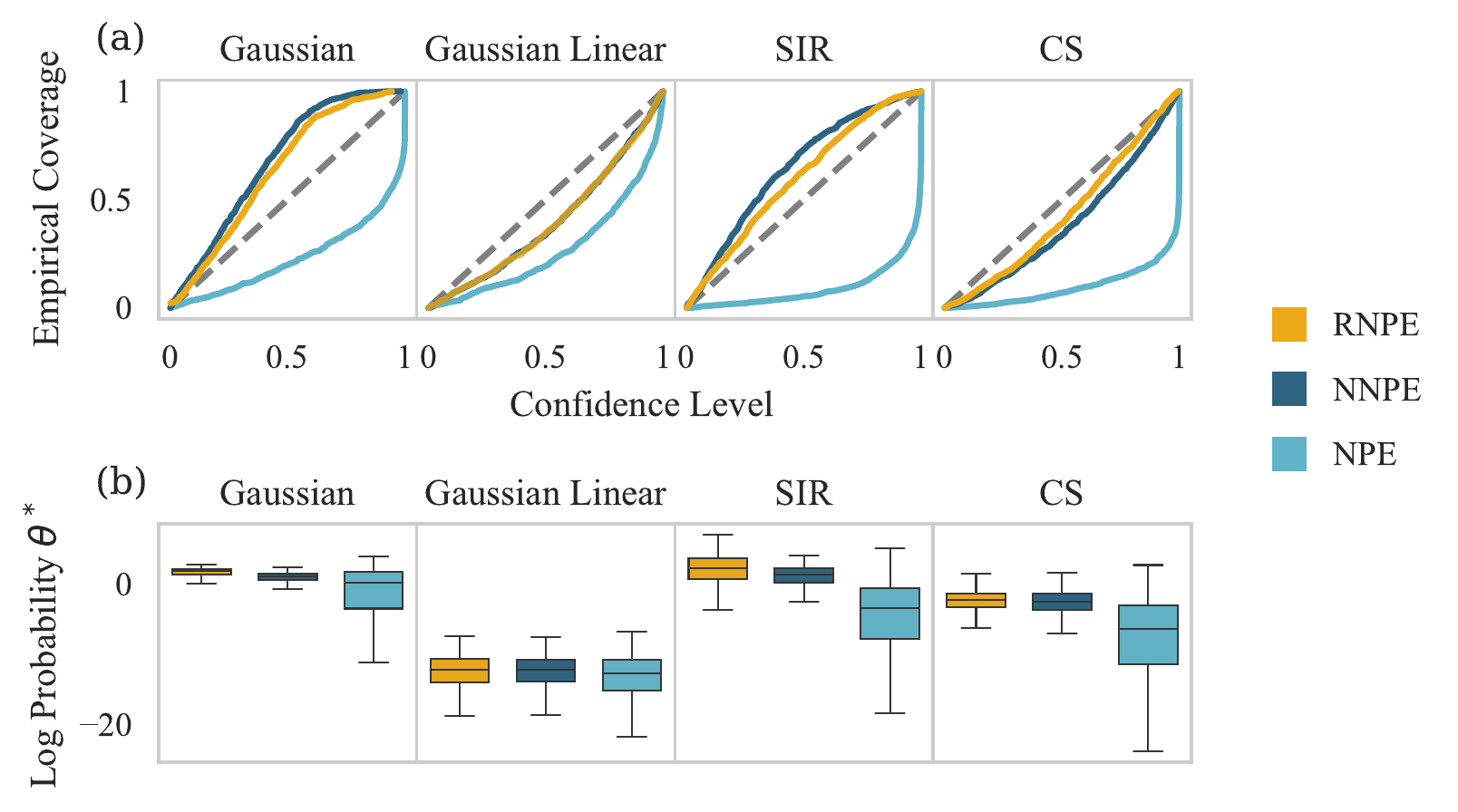}
    \caption{Comparative performance of \gls{rnpe}, \gls{nnpe} and \gls{npe} on the four misspecified tasks, using $1000$ different observation and true parameter pairs. (a) The coverage of each approach at different confidence levels. A well-calibrated posterior follows the dotted line, with conservative posteriors lying above the dotted line, and overconfident posteriors falling below the dotted line. (b) The log-probability of the true parameters $\+\theta^*$ in the approximate posteriors.}
    \label{fig:overall}
\end{figure}

\subsection{Overall Performance}

The results shown in Fig. \ref{fig:overall} show that both \gls{rnpe} and \gls{nnpe} perform better than \gls{npe}, both in terms of producing posteriors with more conservative coverage properties (Fig. \ref{fig:overall}a) and by tending to place more mass on $\+\theta^*$ (Fig. \ref{fig:overall}b). It is important to note that having conservative posterior estimates is generally considered preferable to having overconfident posterior estimates, as the latter could lead to drawing of erroneous scientific conclusions. \Gls{rnpe} slightly outperforms \gls{nnpe} in terms of the mass placed on $\+\theta^*$. We hypothesise this increased performance is due to \gls{rnpe} exploiting the fact that the error model is known and tractable, rather than na\"ively relying on \gls{npe} to account for it. For all the tasks the error model is misspecified: these results highlight the importance of accounting for model misspecification, even if doing so may introduce a misspecified error model.

\subsection{Examples}

\textbf{Gaussian.} An example for the Gaussian task can be seen in Fig. \ref{fig:diagram}, in which $x_1$ corresponds to the mean statistic, and $x_2$ the variance. From observing the posterior misspecification probabilities and the denoised samples, the results clearly identify a discrepancy in which the observed variance statistic is higher than the simulated variance statistics, allowing identification of source of misspecification. As this is a simple example, we can analytically access the true posterior (obtained using the true data generating model). \Gls{rnpe} produces a posterior close to the ground-truth, whereas \gls{npe} confidently places most of the posterior mass far from the ground-truth posterior.

\begin{wrapfigure}[17]{r}{0.5\textwidth}
\includegraphics[width=0.45\textwidth]{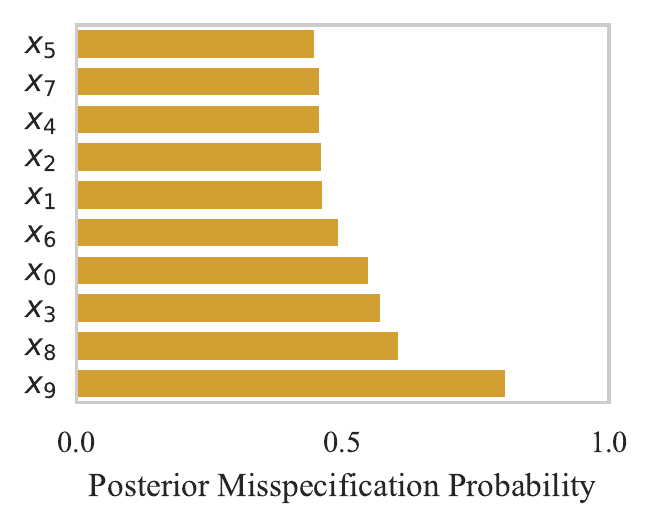}
\caption{Posterior misspecification probabilities for an example of the Gaussian Linear task.}
\label{fig:GaussianLinear_misspecification_probabilities}
\end{wrapfigure} 

\textbf{Gaussian Linear.} For the Gaussian Linear task, misspecification was introduced by applying additive Gaussian noise.  Due to the lack of structure (e.g. correlations between simulation output dimensions), and the lack of sparsity in the errors, the model generally cannot provide much evidence for or against misspecification, as shown by Fig. \ref{fig:GaussianLinear_misspecification_probabilities}. Corresponding pair plots of the denoised data and posterior are shown in Figs. \ref{fig:GaussianLinear_denoised_pairplot}-\ref{fig:GaussianLinear_posterior_pairplot}, in the appendix.

\begin{figure}[h]
    \centering
    \includegraphics[width=\textwidth]{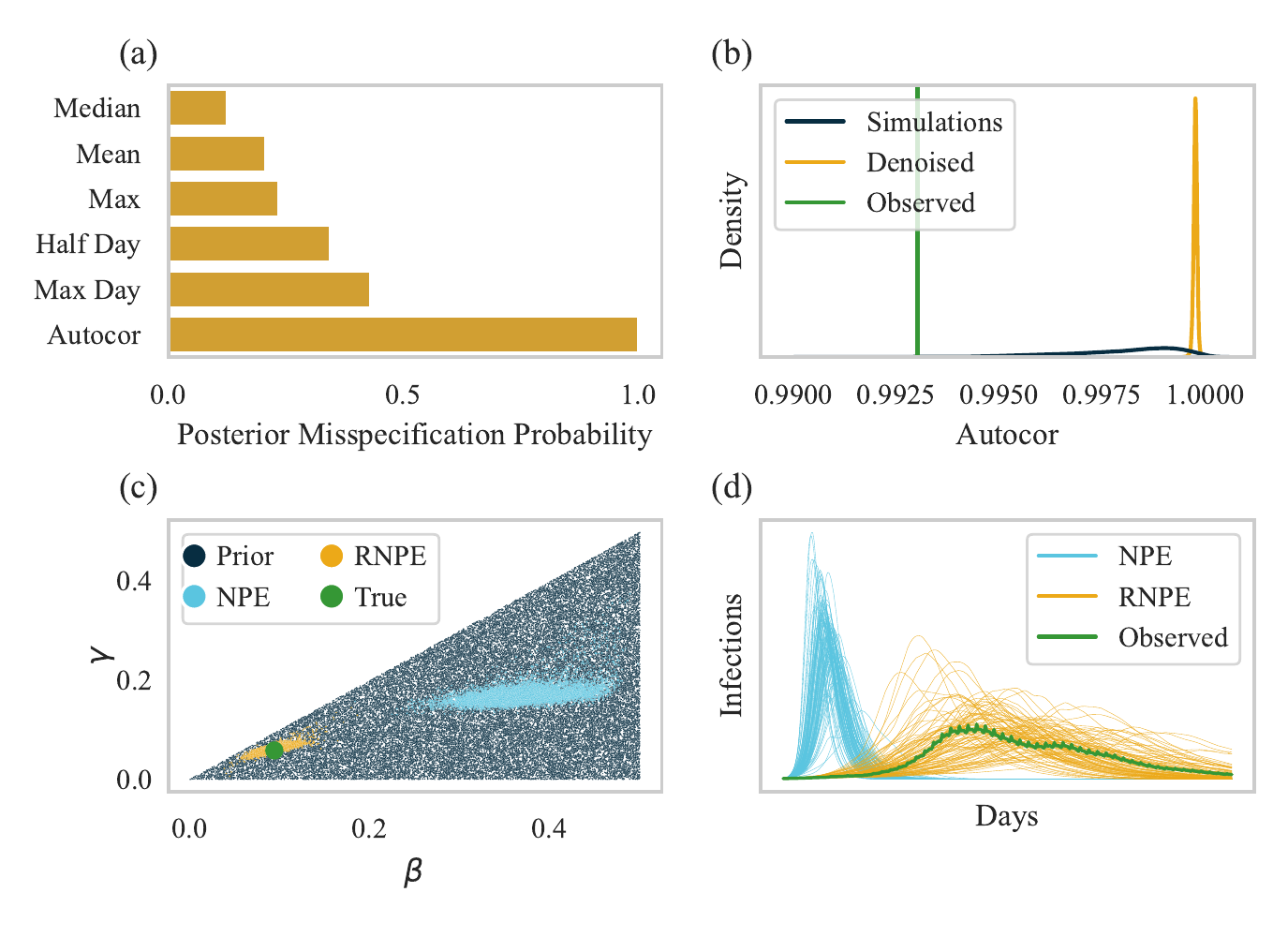}
    \caption{The results for an example of the SIR task. (a) The inferred posterior misspecification probabilities for each summary statistic. (b) Density plot for the denoised samples of the most misspecified summary statistic (autocorrelation). (c) The posteriors for both \gls{npe} and \gls{rnpe} and; (d) The associated posterior predictive distributions for \gls{npe} and \gls{rnpe}. Note the (artificially introduced) reporting delays are visible for the raw observed data.}
    \label{fig:sir_example}
\end{figure}

\textbf{SIR.} In the SIR model, the observed data were artificially corrupted by mimicking reporting delays over the weekend. With this in mind, we can consider how the \emph{model criticism} aspect of \gls{rnpe} could facilitate identification of  the issue. Fig. \ref{fig:sir_example}a shows that most  summary statistics are more likely to be well-specified than was believed \textit{a priori}, with the clear exception of the \texttt{Autocor} summary statistic, which is inferred to be misspecified. Fig. \ref{fig:sir_example}b shows a density plot of the denoised data for the \texttt{Autocor} summary statistic (for a full pair plot, see Fig. \ref{fig:sir_denoised_pairplot}, in the appendix), showing that the observed autocorrelation is much lower than would be expected under the model. This may prompt the practitioner to investigate the day-to-day changes in infections, in which case the discontinuities introduced by the reporting delays could be identified.
 Fig. \ref{fig:sir_example}c shows the posteriors for \gls{rnpe} and \gls{npe}. \Gls{rnpe} yields a posterior with far lower variance than \gls{npe}, demonstrating that it does not always achieve robustness simply by producing broader posterior estimates.

\begin{figure*}
    \centering
\begin{subfigure}[t]{0.5\textwidth}
    \centering
    \includegraphics[width=\textwidth]{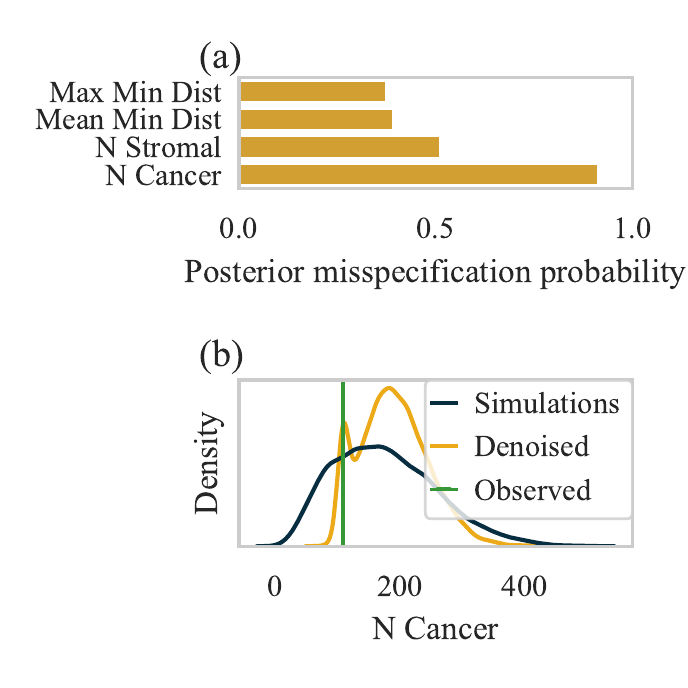}
\end{subfigure}
\begin{subfigure}[t]{0.49\textwidth}
    \centering
    \includegraphics[width=\textwidth]{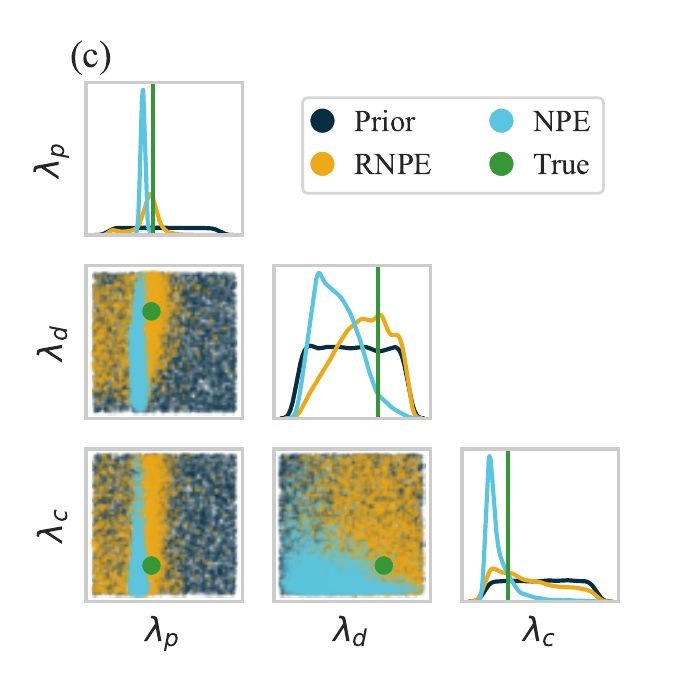}
\end{subfigure}
\caption{The results for an example of the CS task. (a) The inferred posterior misspecification probabilities for each summary statistic. (b) Density plot for the denoised samples of the \texttt{N Cancer} summary statistic. (c) A pair plot of the posteriors for both \gls{npe} and \gls{rnpe}.}
\label{fig:cancer_example}
\end{figure*}

\textbf{CS.} In the CS model, the observed data was corrupted by removing cells in the core regions of tumours, mimicking necrosis (see Fig. \ref{fig:misspecification_cancer} in the appendix). This misspecification primarily impacts the \texttt{N Cancer} summary statistic (the number of cancer cells). From Figs. \ref{fig:cancer_example}a,\ref{fig:cancer_example}b, we can see that the results provide evidence that the \texttt{N Cancer} statistic is misspecified, with the observed data having fewer cancer cells than would be expected if it was generated from the simulator. Once again, alongside domain knowledge, this information points the practitioner to the source of misspecification (necrosis), and hence the model could be adjusted appropriately. For this example, \gls{npe} produced an overconfident posterior with little mass on the true parameters, whereas \gls{rnpe} does not exhibit this issue (Fig. \ref{fig:cancer_example}c).

\subsection{Indeterminacy}
In some cases, there can be indeterminacy when inferring misspecification, in which it is impossible to determine which summary statistic is misspecified, despite clear evidence of misspecification. This occurs, for example, when a model can only be consistent with one observed summary statistic, e.g. $x_1$, at the expense of being able to recreate another summary statistic, e.g. $x_2$ (or vice versa). \Gls{rnpe} allows these trade-offs to be identified, as the statistics will show a small frequency of being jointly well-specified, $\text{Pr}(z_1=0,\ z_2=0 \mid \+y)$, alongside reasonable probabilities of either statistic being well-specified, $\text{Pr}(z_1=1, z_2=0 \mid \+y)$ and $\text{Pr}(z_1=0,\ z_2=1 \mid \+y)$. We show an example of this in Fig. \ref{fig:indeterminacy} in the Appendix.

\section{Conclusions}
\label{section:conclusions}
We described \gls{rnpe}, an approach that can be applied to any black-box simulator to simultaneously perform \emph{model criticism} and \emph{robust inference}. \Gls{rnpe} is philosophically similar to existing robust \gls{sbi} approaches for \gls{abc} and synthetic likelihood \citep{wilkinson2013approximate, frazier2021robust}. However, this work introduces three novel contributions: \emph{i)} we apply the idea of using error models for robust inference to a neural \gls{sbi} method, which has advantages in simulation efficiency, flexibility and scaling to high dimensional problems \citep{lueckmann2021benchmarking}, \emph{ii)} we facilitate model criticism \emph{decoupled} from inference, which avoids confounding of inference failures and model misspecification, and \emph{iii)} we use a spike-and-slab error model to obtain explicit misspecification probability diagnostics for summary statistics, to facilitate \emph{model criticism}. Our work is not without limitations. Firstly, \gls{rnpe} is more computationally costly than \gls{npe}, requiring an additional \gls{mcmc} step and fitting of $q(\+x)$. Secondly, we only evaluated performance of a single choice of error model on a limited number of tasks. It would be beneficial in the future to have a larger set of realistic misspecified tasks for assessing performance, similar to the work by \cite{lueckmann2021benchmarking} for the well-specified case. Finally, we restricted our focus to \gls{npe}; in future work it would be interesting to consider other \gls{sbi} algorithms and to compare results with existing methods for robust inference.

We demonstrated with four examples that \gls{rnpe} consistently outperforms \gls{npe} for parameter inference under misspecification, tending to put more posterior mass on the true parameters whilst also exhibiting more conservative coverage properties. The tasks were designed to mimic levels of misspecification that could be realistically encountered, and hence the poor performance of \gls{npe} calls into question its use in practice, particularly when the level of misspecification could be significant. The substantial difference in performance highlights, as suggested by \cite{box1980sampling}, that we should not have blind faith that a model is sufficiently accurate, and therefore must devise methods for performing \emph{model criticism} and \emph{robust inference}. With further research and wider adoption of these principles in \gls{sbi}, we believe simulators will only become more powerful tools for scientific discovery.

\clearpage
\bibliography{references}

\begin{thebibliography}{}

\bibitem[Atkeson et~al., 2020]{atkeson2020estimating}
Atkeson, A., Kopecky, K., and Zha, T. (2020).
\newblock Estimating and forecasting disease scenarios for {COVID}-19 with an
  {SIR} model.
\newblock Technical report, National Bureau of Economic Research.

\bibitem[Beaumont et~al., 2002]{beaumont2002approximate}
Beaumont, M.~A., Zhang, W., and Balding, D.~J. (2002).
\newblock Approximate {B}ayesian computation in population genetics.
\newblock {\em Genetics}, 162(4):2025--2035.

\bibitem[Bingham et~al., 2019]{bingham2019pyro}
Bingham, E., Chen, J.~P., Jankowiak, M., Obermeyer, F., Pradhan, N.,
  Karaletsos, T., Singh, R., Szerlip, P., Horsfall, P., and Goodman, N.~D.
  (2019).
\newblock Pyro: Deep universal probabilistic programming.
\newblock {\em The Journal of Machine Learning Research}, 20(1):973--978.

\bibitem[Box, 1980]{box1980sampling}
Box, G.~E. (1980).
\newblock Sampling and {Bayes}' inference in scientific modelling and
  robustness.
\newblock {\em Journal of the Royal Statistical Society: Series A (General)},
  143(4):383--404.

\bibitem[Bradbury et~al., 2018]{jax2018github}
Bradbury, J., Frostig, R., Hawkins, P., Johnson, M.~J., Leary, C., Maclaurin,
  D., Necula, G., Paszke, A., Vander{P}las, J., Wanderman-{M}ilne, S., and
  Zhang, Q. (2018).
\newblock {JAX}: composable transformations of {P}ython+{N}um{P}y programs.

\bibitem[Brehmer et~al., 2020]{brehmer2020madminer}
Brehmer, J., Kling, F., Espejo, I., and Cranmer, K. (2020).
\newblock Madminer: Machine learning-based inference for particle physics.
\newblock {\em Computing and Software for Big Science}, 4(1):1--25.

\bibitem[Cannon et~al., 2022]{withcannonyoucan}
Cannon, P., Ward, D., and Schmon, S.~M. (2022).
\newblock Investigating the impact of model misspecification in neural
  simulation-based inference.
\newblock {\em arXiv preprint arXiv:2209.01845}.

\bibitem[Chan et~al., 2018]{chan2018likelihood}
Chan, J., Perrone, V., Spence, J., Jenkins, P., Mathieson, S., and Song, Y.
  (2018).
\newblock A likelihood-free inference framework for population genetic data
  using exchangeable neural networks.
\newblock {\em Advances in neural information processing systems}, 31.

\bibitem[Chen et~al., 2020]{chen2020neural}
Chen, Y., Zhang, D., Gutmann, M., Courville, A., and Zhu, Z. (2020).
\newblock Neural approximate sufficient statistics for implicit models.
\newblock {\em arXiv preprint arXiv:2010.10079}.

\bibitem[Ch{\'e}rief-Abdellatif and Alquier, 2020]{cherief2020mmd}
Ch{\'e}rief-Abdellatif, B.-E. and Alquier, P. (2020).
\newblock {MMD-Bayes}: Robust {Bayesian} estimation via maximum mean
  discrepancy.
\newblock In {\em Symposium on Advances in Approximate Bayesian Inference},
  pages 1--21. PMLR.

\bibitem[Cranmer et~al., 2020]{cranmer2020frontier}
Cranmer, K., Brehmer, J., and Louppe, G. (2020).
\newblock The frontier of simulation-based inference.
\newblock {\em Proceedings of the National Academy of Sciences},
  117(48):30055--30062.

\bibitem[De~Cao et~al., 2020]{de2020block}
De~Cao, N., Aziz, W., and Titov, I. (2020).
\newblock Block neural autoregressive flow.
\newblock In {\em Uncertainty in artificial intelligence}, pages 1263--1273.
  PMLR.

\bibitem[Duane et~al., 1987]{duane1987hybrid}
Duane, S., Kennedy, A.~D., Pendleton, B.~J., and Roweth, D. (1987).
\newblock Hybrid {M}onte {C}arlo.
\newblock {\em Physics letters B}, 195(2):216--222.

\bibitem[Durkan et~al., 2019]{durkan2019neural}
Durkan, C., Bekasov, A., Murray, I., and Papamakarios, G. (2019).
\newblock Neural spline flows.
\newblock {\em Advances in neural information processing systems}, 32.

\bibitem[Durkan et~al., 2020]{durkan2020contrastive}
Durkan, C., Murray, I., and Papamakarios, G. (2020).
\newblock On contrastive learning for likelihood-free inference.
\newblock In {\em International Conference on Machine Learning}, pages
  2771--2781. PMLR.

\bibitem[Dyer et~al., 2021]{dyer2021approximate}
Dyer, J., Cannon, P., and Schmon, S.~M. (2021).
\newblock Approximate bayesian computation with path signatures.
\newblock {\em arXiv preprint arXiv:2106.12555}.

\bibitem[Fearnhead and Prangle, 2012]{fearnhead2012constructing}
Fearnhead, P. and Prangle, D. (2012).
\newblock Constructing summary statistics for approximate {Bayesian}
  computation: semi-automatic approximate {Bayesian} computation.
\newblock {\em Journal of the Royal Statistical Society: Series B (Statistical
  Methodology)}, 74(3):419--474.

\bibitem[Ferguson et~al., 2020]{ferguson2020impact}
Ferguson, N.~M., Laydon, D., Nedjati-Gilani, G., Imai, N., Ainslie, K.,
  Baguelin, M., Bhatia, S., Boonyasiri, A., Cucunub{\'a}, Z., Cuomo-Dannenburg,
  G., et~al. (2020).
\newblock Impact of non-pharmaceutical interventions ({NPI}s) to reduce
  {COVID}-19 mortality and healthcare demand.

\bibitem[Frazier and Drovandi, 2021]{frazier2021robust}
Frazier, D.~T. and Drovandi, C. (2021).
\newblock Robust approximate {Bayesian} inference with synthetic likelihood.
\newblock {\em Journal of Computational and Graphical Statistics},
  30(4):958--976.

\bibitem[Frazier et~al., 2020a]{frazier2020robust}
Frazier, D.~T., Drovandi, C., and Loaiza-Maya, R. (2020a).
\newblock Robust approximate {Bayesian} computation: An adjustment approach.
\newblock {\em arXiv preprint arXiv:2008.04099}.

\bibitem[Frazier et~al., 2020b]{frazier2020model}
Frazier, D.~T., Robert, C.~P., and Rousseau, J. (2020b).
\newblock Model misspecification in approximate {Bayesian} computation:
  consequences and diagnostics.
\newblock {\em Journal of the Royal Statistical Society: Series B (Statistical
  Methodology)}, 82(2):421--444.

\bibitem[Friedman, 2004]{friedman2004multivariate}
Friedman, J. (2004).
\newblock On multivariate goodness-of-fit and two-sample testing.
\newblock Technical report, Citeseer.

\bibitem[Fujisawa et~al., 2021]{fujisawa2021gamma}
Fujisawa, M., Teshima, T., Sato, I., and Sugiyama, M. (2021).
\newblock $\gamma$-abc: Outlier-robust approximate bayesian computation based
  on a robust divergence estimator.
\newblock In {\em International Conference on Artificial Intelligence and
  Statistics}, pages 1783--1791. PMLR.

\bibitem[Gelman and Shalizi, 2013]{gelman2013philosophy}
Gelman, A. and Shalizi, C.~R. (2013).
\newblock Philosophy and the practice of {B}ayesian statistics.
\newblock {\em British Journal of Mathematical and Statistical Psychology},
  66(1):8--38.

\bibitem[George and McCulloch, 1993]{george1993variable}
George, E.~I. and McCulloch, R.~E. (1993).
\newblock Variable selection via {Gibbs} sampling.
\newblock {\em Journal of the American Statistical Association},
  88(423):881--889.

\bibitem[Greenberg et~al., 2019]{greenberg2019automatic}
Greenberg, D., Nonnenmacher, M., and Macke, J. (2019).
\newblock Automatic posterior transformation for likelihood-free inference.
\newblock In {\em International Conference on Machine Learning}, pages
  2404--2414. PMLR.

\bibitem[Gr{\"u}nwald and Langford, 2007]{grunwald2007suboptimal}
Gr{\"u}nwald, P. and Langford, J. (2007).
\newblock Suboptimal behavior of {Bayes} and {MDL} in classification under
  misspecification.
\newblock {\em Machine Learning}, 66(2):119--149.

\bibitem[Gr{\"u}nwald and Van~Ommen, 2017]{grunwald2017inconsistency}
Gr{\"u}nwald, P. and Van~Ommen, T. (2017).
\newblock Inconsistency of {Bayesian} inference for misspecified linear models,
  and a proposal for repairing it.
\newblock {\em Bayesian Analysis}, 12(4):1069--1103.

\bibitem[Hermans et~al., 2020]{hermans2020likelihood}
Hermans, J., Begy, V., and Louppe, G. (2020).
\newblock Likelihood-free mcmc with amortized approximate ratio estimators.
\newblock In {\em International Conference on Machine Learning}, pages
  4239--4248. PMLR.

\bibitem[Hermans et~al., 2021]{hermans2021averting}
Hermans, J., Delaunoy, A., Rozet, F., Wehenkel, A., and Louppe, G. (2021).
\newblock Averting a crisis in simulation-based inference.
\newblock {\em arXiv preprint arXiv:2110.06581}.

\bibitem[Hyndman, 1996]{hyndman1996computing}
Hyndman, R.~J. (1996).
\newblock Computing and graphing highest density regions.
\newblock {\em The American Statistician}, 50(2):120--126.

\bibitem[Jones-Todd et~al., 2019]{jones2019identifying}
Jones-Todd, C.~M., Caie, P., Illian, J.~B., Stevenson, B.~C., Savage, A.,
  Harrison, D.~J., and Bown, J.~L. (2019).
\newblock Identifying prognostic structural features in tissue sections of
  colon cancer patients using point pattern analysis.
\newblock {\em Statistics in medicine}, 38(8):1421--1441.

\bibitem[Kidger and Garcia, 2021]{kidger2021equinox}
Kidger, P. and Garcia, C. (2021).
\newblock Equinox: neural networks in {JAX} via callable pytrees and filtered
  transformations.
\newblock {\em arXiv preprint arXiv:2111.00254}.

\bibitem[Kleijn and van~der Vaart, 2012]{kleijn2012bernstein}
Kleijn, B.~J. and van~der Vaart, A.~W. (2012).
\newblock The {B}ernstein-von-{M}ises theorem under misspecification.
\newblock {\em Electronic Journal of Statistics}, 6:354--381.

\bibitem[Lam et~al., 2015]{lam2015numba}
Lam, S.~K., Pitrou, A., and Seibert, S. (2015).
\newblock Numba: A {LLVM}-based python {JIT} compiler.
\newblock In {\em Proceedings of the Second Workshop on the LLVM Compiler
  Infrastructure in HPC}, pages 1--6.

\bibitem[Lopez-Paz and Oquab, 2016]{lopez2016revisiting}
Lopez-Paz, D. and Oquab, M. (2016).
\newblock Revisiting classifier two-sample tests.
\newblock {\em arXiv preprint arXiv:1610.06545}.

\bibitem[Lueckmann et~al., 2021]{lueckmann2021benchmarking}
Lueckmann, J.-M., Boelts, J., Greenberg, D., Goncalves, P., and Macke, J.
  (2021).
\newblock Benchmarking simulation-based inference.
\newblock In {\em International Conference on Artificial Intelligence and
  Statistics}, pages 343--351. PMLR.

\bibitem[Lueckmann et~al., 2017]{lueckmann2017flexible}
Lueckmann, J.-M., Goncalves, P.~J., Bassetto, G., {\"O}cal, K., Nonnenmacher,
  M., and Macke, J.~H. (2017).
\newblock Flexible statistical inference for mechanistic models of neural
  dynamics.
\newblock {\em Advances in neural information processing systems}, 30.

\bibitem[Metropolis et~al., 1953]{metropolis1953equation}
Metropolis, N., Rosenbluth, A.~W., Rosenbluth, M.~N., Teller, A.~H., and
  Teller, E. (1953).
\newblock Equation of state calculations by fast computing machines.
\newblock {\em The journal of chemical physics}, 21(6):1087--1092.

\bibitem[Miglino et~al., 1995]{miglino1995evolving}
Miglino, O., Lund, H.~H., and Nolfi, S. (1995).
\newblock Evolving mobile robots in simulated and real environments.
\newblock {\em Artificial life}, 2(4):417--434.

\bibitem[Neal et~al., 2011]{neal2011mcmc}
Neal, R.~M. et~al. (2011).
\newblock {MCMC} using {H}amiltonian dynamics.
\newblock {\em Handbook of markov chain monte carlo}, 2(11):2.

\bibitem[Papamakarios and Murray, 2016]{papamakarios2016fast}
Papamakarios, G. and Murray, I. (2016).
\newblock Fast $\varepsilon$-free inference of simulation models with
  {Bayesian} conditional density estimation.
\newblock {\em Advances in neural information processing systems}, 29.

\bibitem[Papamakarios et~al., 2019a]{papamakarios2019normalizing}
Papamakarios, G., Nalisnick, E., Rezende, D.~J., Mohamed, S., and
  Lakshminarayanan, B. (2019a).
\newblock Normalizing flows for probabilistic modeling and inference.
\newblock {\em arXiv preprint arXiv:1912.02762}.

\bibitem[Papamakarios et~al., 2019b]{papamakarios2019sequential}
Papamakarios, G., Sterratt, D., and Murray, I. (2019b).
\newblock Sequential neural likelihood: Fast likelihood-free inference with
  autoregressive flows.
\newblock In {\em The 22nd International Conference on Artificial Intelligence
  and Statistics}, pages 837--848. PMLR.

\bibitem[Park et~al., 2015]{park2015k2}
Park, M., Jitkrittum, W., and Sejdinovic, D. (2015).
\newblock {K2-ABC}: approximate {Bayesian} computation with kernel embeddings.
\newblock {\em arXiv preprint arXiv:1502.02558}.

\bibitem[Perla et~al., 2022]{perla2022quantitative}
Perla, J., Sargent, T.~J., Stachurski, J., and Rackauckas, C. (2022).
\newblock Quantitative economics with julia: Modeling shocks in {COVID}-19 with
  stochastic differential equations.

\bibitem[Phan et~al., 2019]{phan2019composable}
Phan, D., Pradhan, N., and Jankowiak, M. (2019).
\newblock Composable effects for flexible and accelerated probabilistic
  programming in {N}um{P}yro.
\newblock {\em arXiv preprint arXiv:1912.11554}.

\bibitem[Prangle et~al., 2014]{prangle2014diagnostic}
Prangle, D., Blum, M.~G., Popovic, G., and Sisson, S. (2014).
\newblock Diagnostic tools for approximate {Bayesian} computation using the
  coverage property.
\newblock {\em Australian \& New Zealand Journal of Statistics},
  56(4):309--329.

\bibitem[Pritchard et~al., 1999]{pritchard1999population}
Pritchard, J.~K., Seielstad, M.~T., Perez-Lezaun, A., and Feldman, M.~W.
  (1999).
\newblock Population growth of human y chromosomes: a study of y chromosome
  microsatellites.
\newblock {\em Molecular biology and evolution}, 16(12):1791--1798.

\bibitem[Rackauckas and Nie, 2017]{rackauckas2017differentialequations}
Rackauckas, C. and Nie, Q. (2017).
\newblock {D}ifferential{E}quations. jl--a performant and feature-rich
  ecosystem for solving differential equations in julia.
\newblock {\em Journal of Open Research Software}, 5(1).

\bibitem[Randall et~al., 2007]{randall2007climate}
Randall, D.~A., Wood, R.~A., Bony, S., Colman, R., Fichefet, T., Fyfe, J.,
  Kattsov, V., Pitman, A., Shukla, J., Srinivasan, J., et~al. (2007).
\newblock Climate models and their evaluation.
\newblock In {\em Climate change 2007: The physical science basis. Contribution
  of Working Group I to the Fourth Assessment Report of the IPCC (FAR)}, pages
  589--662. Cambridge University Press.

\bibitem[Schmitt et~al., 2021]{schmitt2021bayesflow}
Schmitt, M., B{\"u}rkner, P.-C., K{\"o}the, U., and Radev, S.~T. (2021).
\newblock {B}ayes{F}low can reliably detect model misspecification and
  posterior errors in amortized {B}ayesian inference.
\newblock {\em arXiv preprint arXiv:2112.08866}.

\bibitem[Schmon et~al., 2020]{schmon2020generalized}
Schmon, S.~M., Cannon, P.~W., and Knoblauch, J. (2020).
\newblock Generalized posteriors in approximate {Bayesian} computation.
\newblock {\em arXiv preprint arXiv:2011.08644}.

\bibitem[Syring and Martin, 2019]{syring2019calibrating}
Syring, N. and Martin, R. (2019).
\newblock Calibrating general posterior credible regions.
\newblock {\em Biometrika}, 106(2):479--486.

\bibitem[Talbott, 2016]{sep-epistemology-bayesian}
Talbott, W. (2016).
\newblock {Bayesian Epistemology}.
\newblock In Zalta, E.~N., editor, {\em The {Stanford} Encyclopedia of
  Philosophy}. Metaphysics Research Lab, Stanford University, {W}inter 2016
  edition.

\bibitem[Talts et~al., 2018]{talts2018validating}
Talts, S., Betancourt, M., Simpson, D., Vehtari, A., and Gelman, A. (2018).
\newblock Validating {Bayesian} inference algorithms with simulation-based
  calibration.
\newblock {\em arXiv preprint arXiv:1804.06788}.

\bibitem[Tavar{\'e} et~al., 1997]{tavare1997inferring}
Tavar{\'e}, S., Balding, D.~J., Griffiths, R.~C., and Donnelly, P. (1997).
\newblock Inferring coalescence times from {DNA} sequence data.
\newblock {\em Genetics}, 145(2):505--518.

\bibitem[Thomas et~al., 2022]{thomas2022likelihood}
Thomas, O., Dutta, R., Corander, J., Kaski, S., and Gutmann, M.~U. (2022).
\newblock Likelihood-free inference by ratio estimation.
\newblock {\em Bayesian Analysis}, 17(1):1--31.

\bibitem[Wilkinson, 2013]{wilkinson2013approximate}
Wilkinson, R.~D. (2013).
\newblock Approximate {B}ayesian computation ({ABC}) gives exact results under
  the assumption of model error.
\newblock {\em Statistical applications in genetics and molecular biology},
  12(2):129--141.

\bibitem[Wood, 2010]{wood2010statistical}
Wood, S.~N. (2010).
\newblock Statistical inference for noisy nonlinear ecological dynamic systems.
\newblock {\em Nature}, 466(7310):1102--1104.

\bibitem[Zhou, 2020]{zhou2020mixed}
Zhou, G. (2020).
\newblock Mixed {H}amiltonian {M}onte {C}arlo for mixed discrete and continuous
  variables.
\newblock {\em Advances in Neural Information Processing Systems},
  33:17094--17104.

\bibitem[Zhou, 2022]{zhou2022metropolis}
Zhou, G. (2022).
\newblock Metropolis augmented {H}amiltonian {M}onte {C}arlo.
\newblock {\em arXiv preprint arXiv:2201.08044}.

\end{thebibliography}

\clearpage
\appendix
\section*{Appendix}

\section{Additional Results}

\label{appendix:additional_results}
\subsection*{C2ST}
\begin{wrapfigure}{r}{0.5\textwidth}
\includegraphics[width=0.4\textwidth]{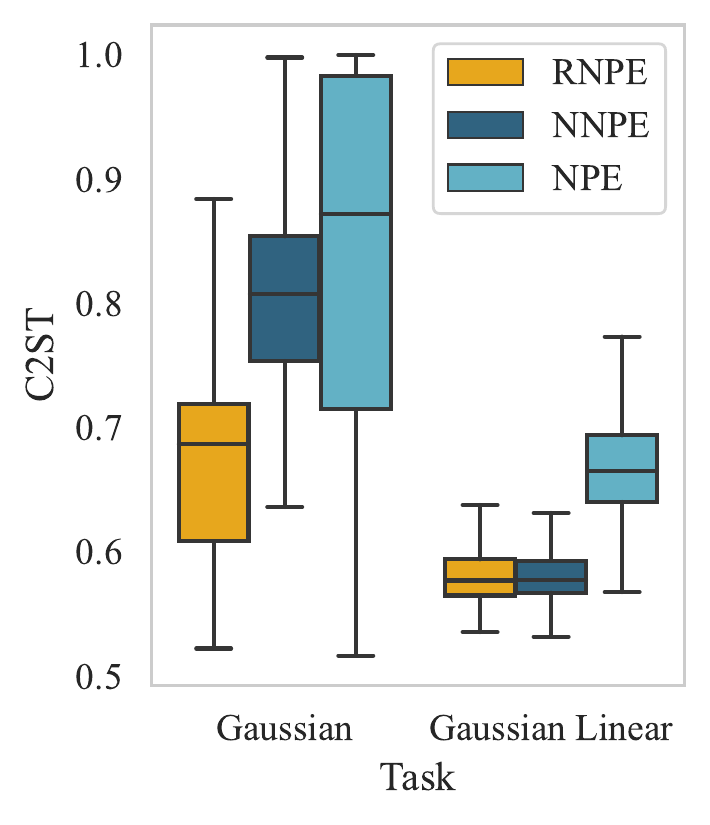}
\caption{The C2ST accuracy, between the true posterior (under the true data generating process), and the approximate posteriors for each method.}
\label{fig:c2st}
\end{wrapfigure} 

In the classifier 2-sample test \citep{friedman2004multivariate, lopez2016revisiting} a classifier is trained to distinguish between a true and inferred posterior distribution. In the case that the classifier struggles to distinguish between the samples (i.e. when the posterior approximation is good), the C2ST score approaches $0.5$. In contrast, when the posterior approximation is poor the C2ST score will be close to $1$. Here we take the \enquote*{true} posterior to be the analytical posterior obtained under the true data generating process. For calculation of the C2ST metric we used a multilayer perceptron, as implemented in the sbibm python package \citep{lueckmann2021benchmarking}. This metric requires a tractable posterior, so is only available for the Gaussian and Gaussian Linear tasks. We note that the error model used in \gls{rnpe} and \gls{nnpe} is misspecified, so we cannot realistically expect a C2ST score close to $0.5$, but the results are useful to highlight that even a misspecified error model can improve performance. These results are shown in Fig. \ref{fig:c2st}

\subsection*{Mean Square Error}
In addition to calculating the log-probability of $\+\theta^*$ and the posterior coverage properties (see Section \ref{section:metrics} and Fig.~\ref{fig:overall}), we also considered taking the posterior mean as a parameter point estimate and comparing this to the true parameters  $\+\theta^*$. Over the $1000$ different true parameters and observation pairs, we compared the posterior means to the true parameters using the mean squared error
$$
\operatorname{MSE} = \frac{1}{1000} \sum_{i=1}^{1000} (\bar{\theta}_{ij} - \theta^*_{ij})^2,
$$
where $\bar{\theta}_{ij}$ is the posterior mean of the $j$-th parameter, resulting from either \gls{npe} or \gls{rnpe}, and $\theta^*_{ij}$ is the ground truth parameter sampled from the prior. These results are shown in Table \ref{table:mse}.

\begin{table}[htbp]
\centering
\begin{tabular}{@{}llccc@{}}
\toprule
Task & Parameter & \multicolumn{3}{c}{Mean Squared Error} \\
\cmidrule{3-5}
     &    & NNPE            & NPE  & RNPE\\ \midrule
Gaussian & $\mu$       & 0.04 & 0.17 & 0.02 \\ \addlinespace
Gaussian Linear & $\+\theta$  & 0.70 & 0.75 & 0.70 \\ \addlinespace
CS       & $\lambda_c$ & 0.12 & 0.12 & 0.13 \\
         & $\lambda_d$ & 1.21 & 1.92 & 1.21 \\
         & $\lambda_p$ & 0.80 & 1.10 & 0.75 \\ \addlinespace
SIR      & $\beta$     & 0.20 & 1.10 & 0.13 \\
         & $\gamma$    & 0.11 & 0.59 & 0.09 \\ \bottomrule
         \\
\end{tabular}
\caption{The mean squared error between the posterior means and the true parameters for each task, estimated using $1000$ different observation and true parameter pairs. The mean squared errors were calculated on standardised parameters, i.e. scaled based on the prior mean and variance, due to the differing scales of different parameters. For the Gaussian Linear task, we average the mean squared error across the parameters (rather than listing them for the 10 parameters individually).}
\label{table:mse}
\end{table}

\FloatBarrier
\subsection*{Well-Specified Case}
\begin{figure}[h]
    \centering
    \includegraphics[width=0.7\linewidth]{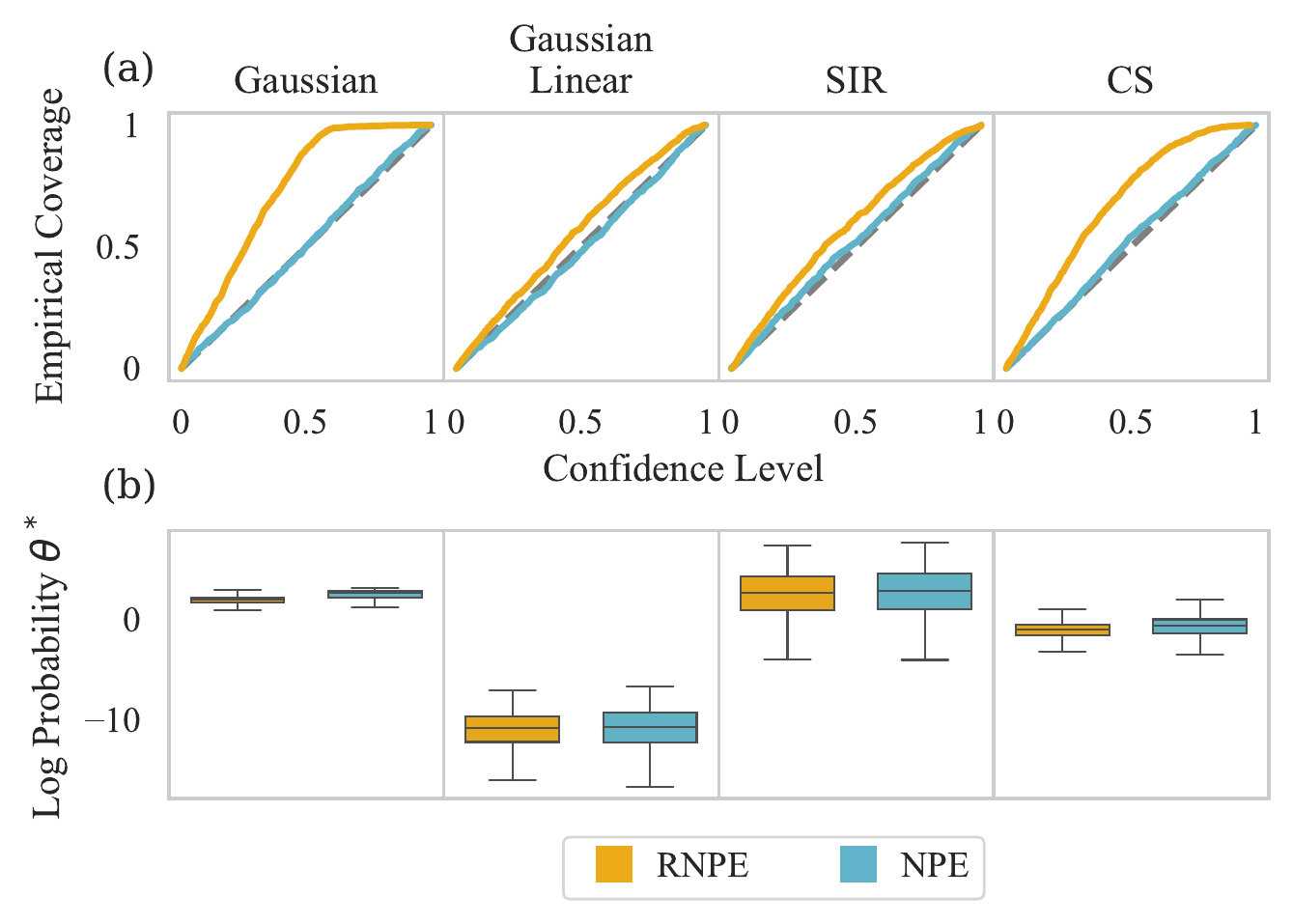}
    \caption{Comparative performance of \gls{rnpe} and \gls{npe} on the four tasks without misspecification of the simulator, using $1000$ different observation and true parameter pairs. (a) The coverage of each approach at different confidence levels. A well-calibrated posterior follows the dotted line, with conservative posteriors lying above the dotted line, and overconfident posteriors falling below the dotted line. (b) The log-probability of the true parameters $\+\theta^*$ in the approximate posteriors.}
    \label{fig:overall_well_specified}
\end{figure}

An obvious question to ask is, what is the impact of incorrectly using an error model when the simulator is well-specified? These results are shown in Fig. \ref{fig:overall_well_specified}. In the well-specified case, \gls{npe} becomes well-calibrated (Fig. \ref{fig:overall_well_specified}a), whereas \gls{rnpe} produces slightly conservative posteriors. \gls{npe} performs marginally better in terms of the posterior mass placed on the true parameters (Fig. \ref{fig:overall_well_specified}b). However, in the well-specified case, \gls{rnpe} may indicate that the model is not likely to be misspecified, in which case the researcher could choose to collapse the error model by directly using $q(\+\theta \mid \+x)$, with no additional cost.

\section{Tasks}
\label{appendix:tasks}

\subsection*{Gaussian}
\setlength{\leftskip}{0.6cm}
\textbf{Model Description.} A simple Gaussian toy model used by \cite{frazier2020model}. The task has a single parameter $\mu$, which is the mean of a Gaussian distribution. A simulation consists of drawing 100 i.i.d. samples from $N(\mu, 1)$, which are summarised to the sample mean and variance.

\textbf{Misspecification.} To artificially introduce misspecification, samples were drawn from $N(\mu, 2)$ for the observed data, rather than $N(\mu, 1)$,  prior to summarising.

\textbf{Priors.} $\mu \sim N(0, 25)$. \\

\setlength{\leftskip}{0cm}
\subsection*{Gaussian Linear}
\setlength{\leftskip}{0.6cm}
\textbf{Model Description.} A misspecified extension of a simple Gaussian toy model used by \cite{lueckmann2021benchmarking}. The task involves inferring the mean vector $\+\theta$ of a 10-d Gaussian distribution, where a simulation consists of sampling this distribution i.e. $\+x \mid \+\theta \sim N(\+\theta,\ 0.1 \cdot \+I_{10})$. 

\textbf{Misspecification.} To artificially introduce misspecification, we add Gaussian additive noise such that $\+y \mid \+x \sim N(\+x,\ 0.1 \cdot \+I_{10})$, or equivalently, we can view this as a misspecified variance in the simulator such that $\+y \mid \+\theta \sim N(\+\theta,\ 0.2 \cdot \+I_{10})$.

\textbf{Priors.} $\+\theta \sim N(0,\ 0.1 \cdot \+I_{10})$. \\

\setlength{\leftskip}{0cm}
\subsection*{SIR}
\setlength{\leftskip}{0.6cm}

\textbf{Model Description.} The SIR model is an idealised model of disease spread, where a series of differential equations describe the rates of transitions between three states: susceptible ($s$), infected ($i$) and recovered ($r$). The simplest (and deterministic) model is
\begin{align}
    \frac{d s}{d t} = -\beta s i, \quad \frac{d i}{d t} = \beta s i - \gamma i, \quad \frac{d r}{d t} = \gamma i,
\end{align}
where $\beta$ is the infection rate, and $\gamma$ the recovery rate. We use a stochastic extension of this model, with a time-varying infection rate $\tilde{\beta}_t$ \citep[see][]{atkeson2020estimating, perla2022quantitative}, reflecting for example policy changes or mutations in the virus. This process is parameterised using the basic reproduction number, $R_{0t} = \frac{\tilde{\beta}_t}{\gamma}$
\begin{align}
dR_{0t} = \eta \left(\frac{\beta}{\gamma} - R_{0t} \right) dt + \sigma \sqrt{R_0}dW_t,
\end{align}
where $\eta$ is the mean reversion strength of $R_{0t}$ to $\frac{\beta}{\gamma}$, $\sigma$ is the volatility, and $W_t$ is Brownian motion. We set $\eta = 0.05$ and $\sigma = 0.05$ and focused on inferring $\beta$ and $\gamma$. This model was implemented in Julia, using the DifferentialEquations package \citep{rackauckas2017differentialequations}. We scaled the resulting infection trajectories by 100,000 (representing the population size), and used summary statistics of the daily infection counts over $365$ days, as described in Section \ref{section:tasks}.

\textbf{Misspecification.} To artificially introduce misspecification, we reduce the infection counts on weekends by 5\%, and these infections are recouped on the subsequent Monday, mimicking reporting delays over weekends. We (arbitrarily) take the first day to be Monday. An example of a simulations, compared to an observation with reporting delays is shown in Fig.~\ref{fig:sir_example}. 

\textbf{Priors.} We used $\beta \sim \text{Uniform}(0, 0.5)$ and $\gamma \sim \text{Uniform}(0, 0.5)$, but rejected samples where $\gamma > \beta$ as frequently the number of infections stayed near zero (as the recovery rate is greater than the infection rate). \\

\setlength{\leftskip}{0cm}
\subsection*{CS}
\setlength{\leftskip}{0.6cm}

\textbf{Model Description.} The CS model simulates cancer and stromal cell development on a 2D plane. The total number of cells $N^c$, number of unobserved parents $N^p$, and the number of daughter cells for each parent $N_i^d$ are sampled
\begin{align*}
    N^c &\sim \text{Poisson}(\lambda^c), \\
    N^p &\sim \text{Poisson}(\lambda^p), \\
    N_i^d &\sim \text{Poisson}(\lambda^d), \quad i=1, \ldots ,N^p,
\end{align*}
where $\lambda^c$, $\lambda^p$ and $\lambda^d$ are the three Poisson rate parameters we attempt to infer. The cell positions $\{c_i\}_{i=1}^{N^c}$ and parent positions $\{p_i\}_{i=1}^{N^p}$ are generated uniformly on the 2D plane (i.e. as two homogeneous point processes). Let $r_i$ represent the euclidean distance from parent $p_i$ to its $N_i^d$-th nearest cell. $r_i$ is the affected radius, such that cells are assigned to be cancerous if they fell on or within this distance of $p_i$. We used summary statistics based on cell type counts and distances, as described in Section \ref{section:tasks}. In order to efficiently calculate the distances, we made use of the Numba just-in-time compiler for Python \citep{lam2015numba}. The distance based metrics (\texttt{Mean Min Dist} and \texttt{Max Min Dist}) were approximated empirically through sampling 50 stromal cells, as the full distance matrix was too expensive to compute.

\begin{figure}
    \centering
    \includegraphics[width=\textwidth]{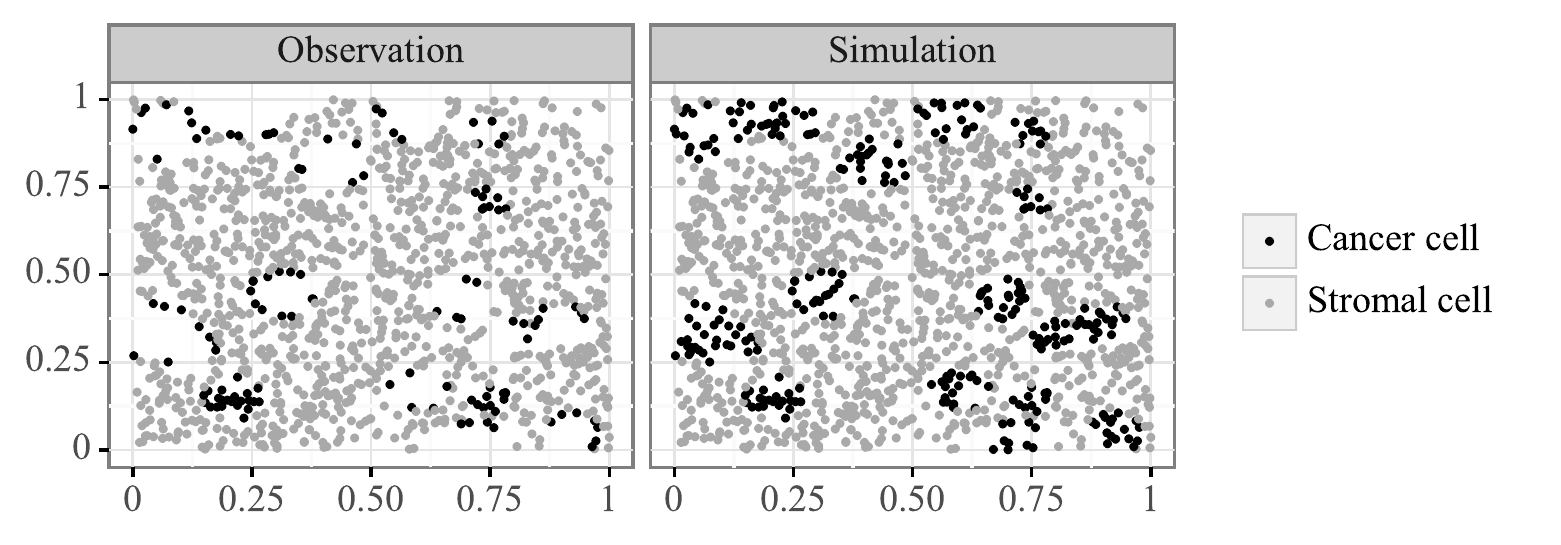}
    \caption{Example of a raw observation for the CS task, compared to the underlying simulation prior to corruption. In the core region of some tumours for the observed data, cancer cells have been removed, artificially mimicking necrosis. See Appendix \ref{appendix:tasks} for more information.}
    \label{fig:misspecification_cancer}
\end{figure}

\textbf{Misspecification.} For each parent, we sampled $w_i \sim \text{Bernoulli}(0.75)$. If $w_i=1$, cancer cells falling within $0.8r_i$ of the corresponding parent $p_i$ were removed. An example of this is shown in Fig.~\ref{fig:misspecification_cancer}. This mimics necrosis that often occurs in the centre regions of tumours \cite[see e.g., ][]{jones2019identifying}. 

\textbf{Priors.} $\lambda_c \sim \text{Uniform}(200, 1500)$, \ $\lambda_p \sim \text{Uniform}(3, 20)$, \ $\lambda_d \sim \text{Uniform}(10, 20).$

\setlength{\leftskip}{0cm}

\section{Hyperparameters}
\label{appendix:hyperparameters}
\textbf{Flows.} For $q(\+x)$ we used a block neural autoregressive flow \citep{de2020block}, with a single hidden layer of size $8D$, where $x\in \mathbb{R}^D$. For $q(\+\theta \mid \+x)$ we used a neural spline flow \citep{durkan2019neural}, defining the transform on the interval $[-5, 5]$, using 10 spline segments, and 5 coupling layers. For both flows we used a standard Gaussian base distribution. We used custom normalising flow implementations using JAX \citep{jax2018github} and Equinox \citep{kidger2021equinox}, which can be found at \url{https://github.com/danielward27/flowjax}. We found JAX flow implementations alongside NumPyro \citep{phan2019composable} for performing \gls{hmc} to be several orders of magnitude faster than relying on Pytorch implementations and Pyro \citep{bingham2019pyro}.

\textbf{Training.} For both flow approximations, we used a batch size of 256, using 10\% of the data as a validation set. We trained for a maximum of 50 epochs, terminating training early if 5 consecutive epochs occurred with no improvement in the validation loss. For the block neural autoregressive flow $q(\+x)$, we used a learning rate of $1\times10^{-2}$, whereas for the neural spline flow $q(\+\theta \mid \+x)$ we used a learning rate of $5\times10^{-4}$.

\textbf{\Gls{mcmc}.} We used 100,000 steps with 20,000 warm up steps. The trajectory length was set to 1, and the target acceptance probability to 0.95 to increase the robustness of the \gls{mcmc} algorithm. All other hyperparameters were left to the defaults of the NumPyro Python package \citep{phan2019composable}.

\section{Computational Resources}
All experiments were performed on CPU, with a maximum of 8GB of RAM, using the computational facilities of the Advanced Computing Research Centre, University of Bristol - \url{http://www.bristol.ac.uk/acrc/}.

\FloatBarrier
\section{Additional Figures}
\label{appendix:additional figures}

\begin{figure}[h]
    \centering
    \includegraphics[width=\linewidth]{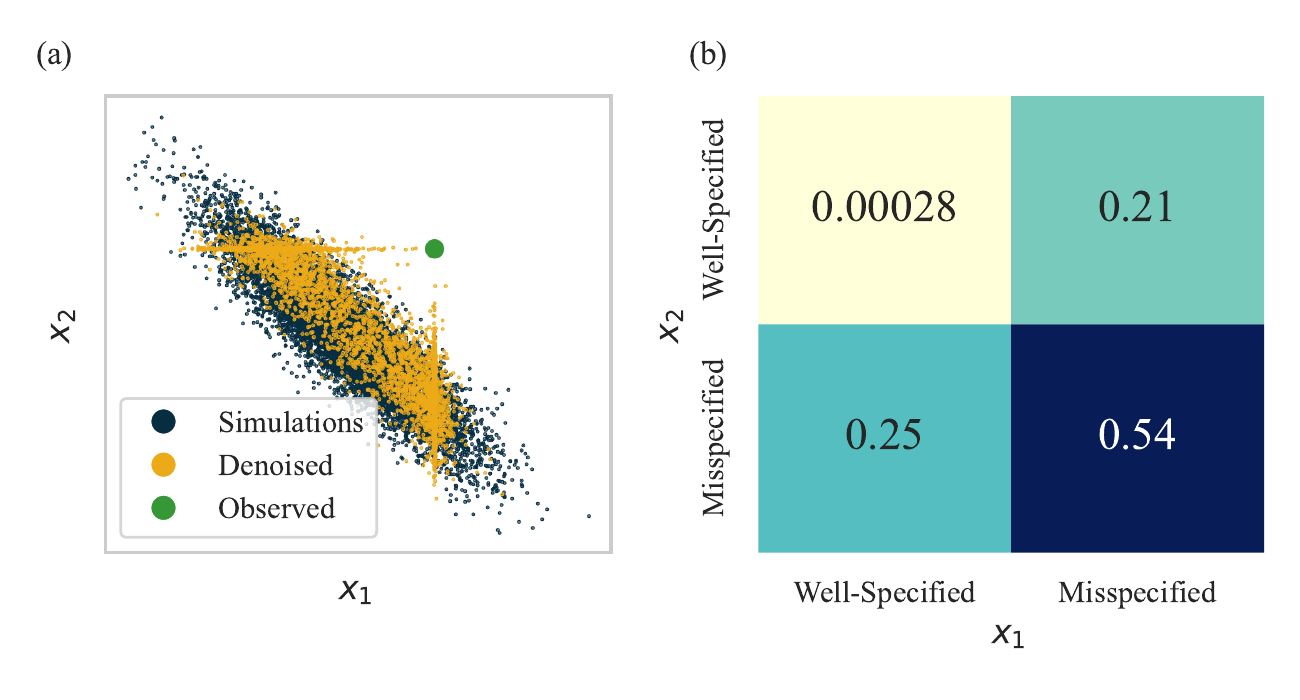}
    \caption{An example in which there is \enquote*{indeterminacy} in the inferred misspecification, where we take $p(\+x)$ to be a correlated Gaussian. (a) The denoised samples, compared to the observed data and simulations. (b) The frequencies with which the different pairwise combinations occurred in $\+z \sim \hat{p}(\+z|\+y)$, inferred during denoising. The results suggest a very low probability of both being well specified, despite reasonable probabilities of either $x_1$ or $x_2$ (exclusively) being well-specified, or both being misspecified. Despite the strong evidence for misspecification, the source of misspecification cannot be determined without further information.}
    \label{fig:indeterminacy}
\end{figure}

\begin{figure}
    \centering
    \includegraphics[width=\textwidth]{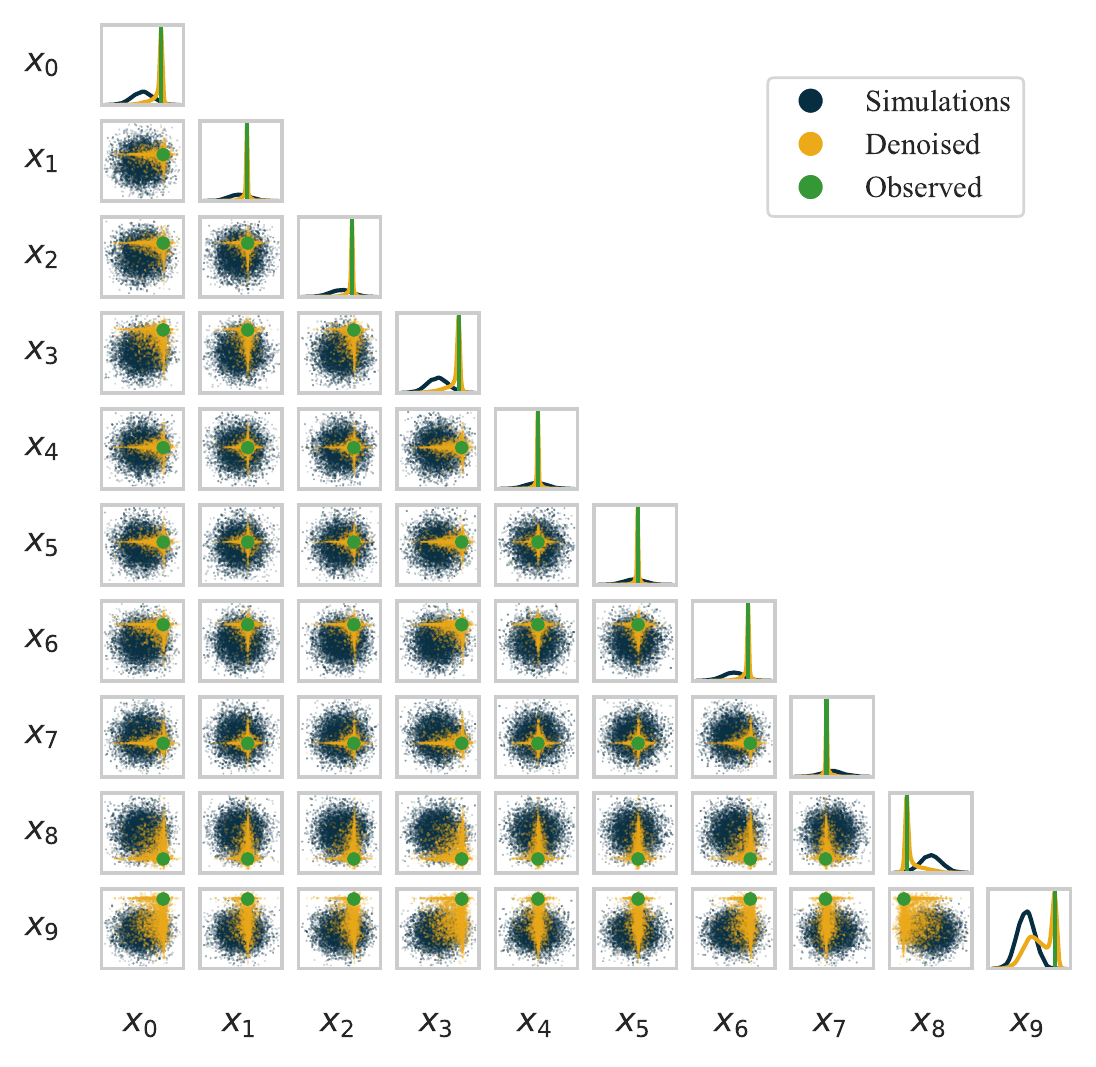}
        \caption{Pair plot of the denoised data for the Gaussian Linear task, corresponding to the example shown in Section \ref{section:results}.}
        \label{fig:GaussianLinear_denoised_pairplot}
\end{figure}

\begin{figure}
    \centering
    \includegraphics[width=\textwidth, ]{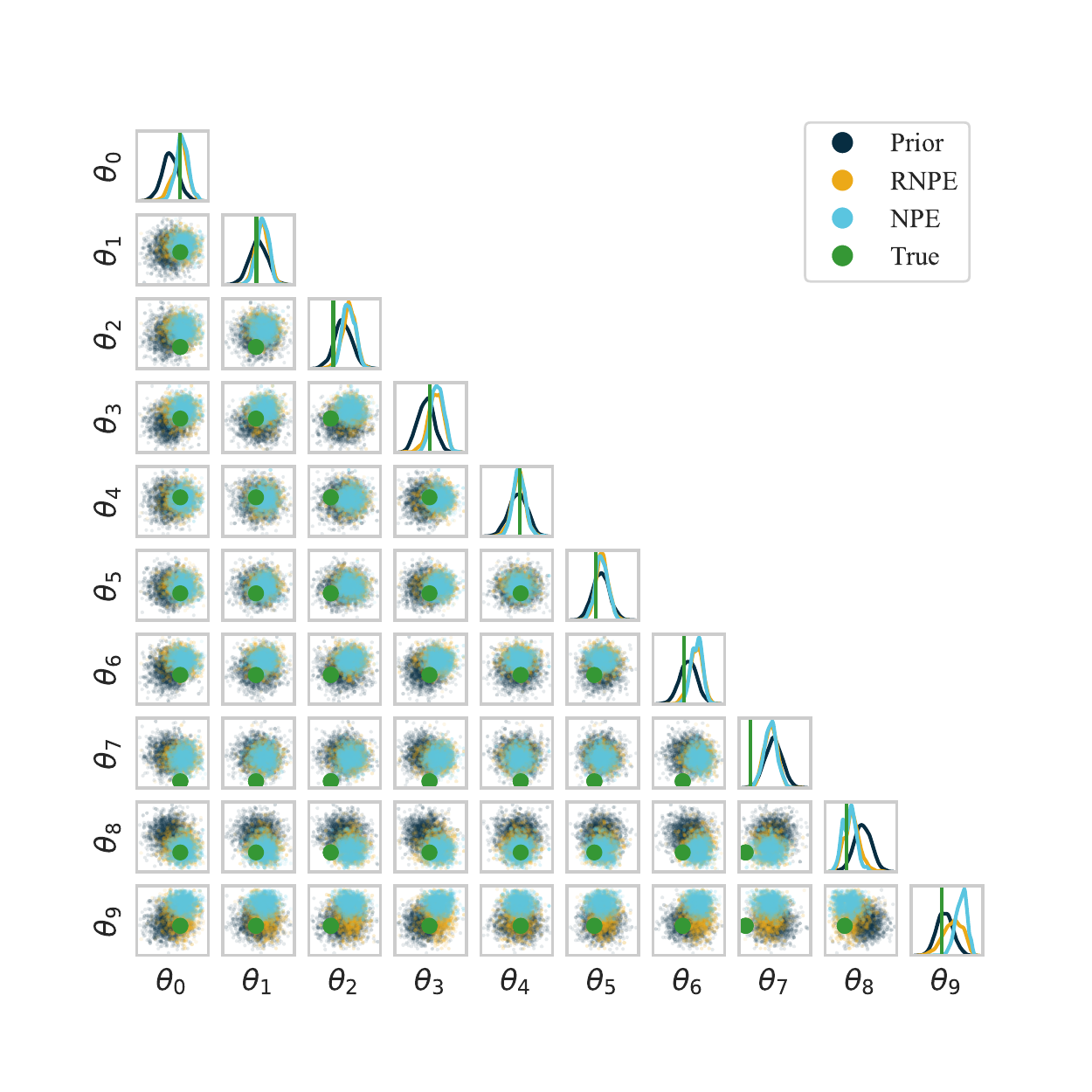}
        \caption{Pair plot of the posterior Gaussian Linear task, corresponding to the example shown in Section \ref{section:results}.}
        \label{fig:GaussianLinear_posterior_pairplot}
\end{figure}

\begin{figure}
    \centering
    \includegraphics{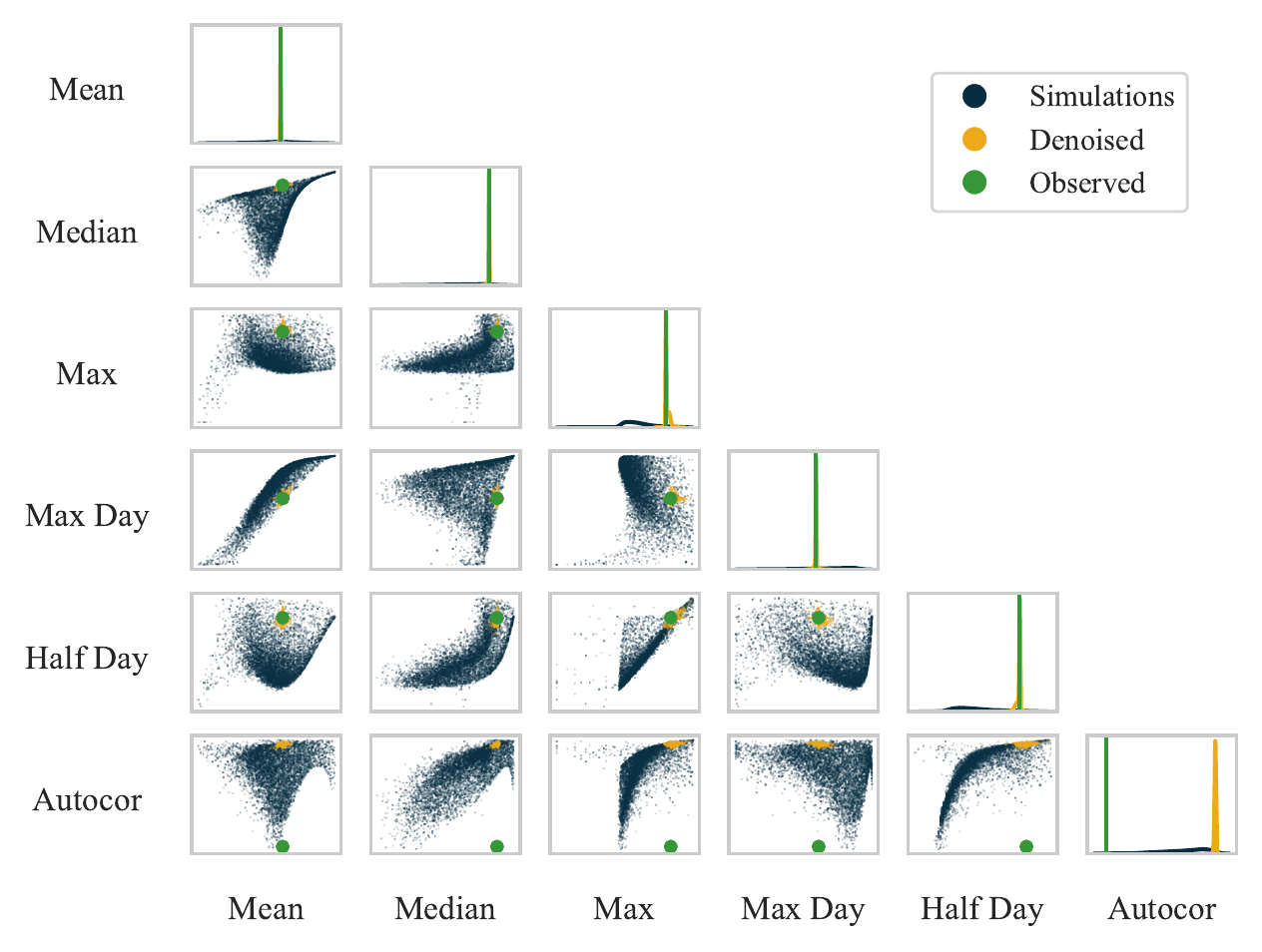}
        \caption{Pair plot of the denoised data for the SIR task, corresponding to the example shown in Section \ref{section:results}. Note that the observed data point obscures the denoised data along many axes, due to the shrinkage effect of the spike and slab error model. See Section \ref{section:tasks} for the summary statistic definitions.}
        \label{fig:sir_denoised_pairplot}
\end{figure}

\begin{figure}
    \centering
    \includegraphics{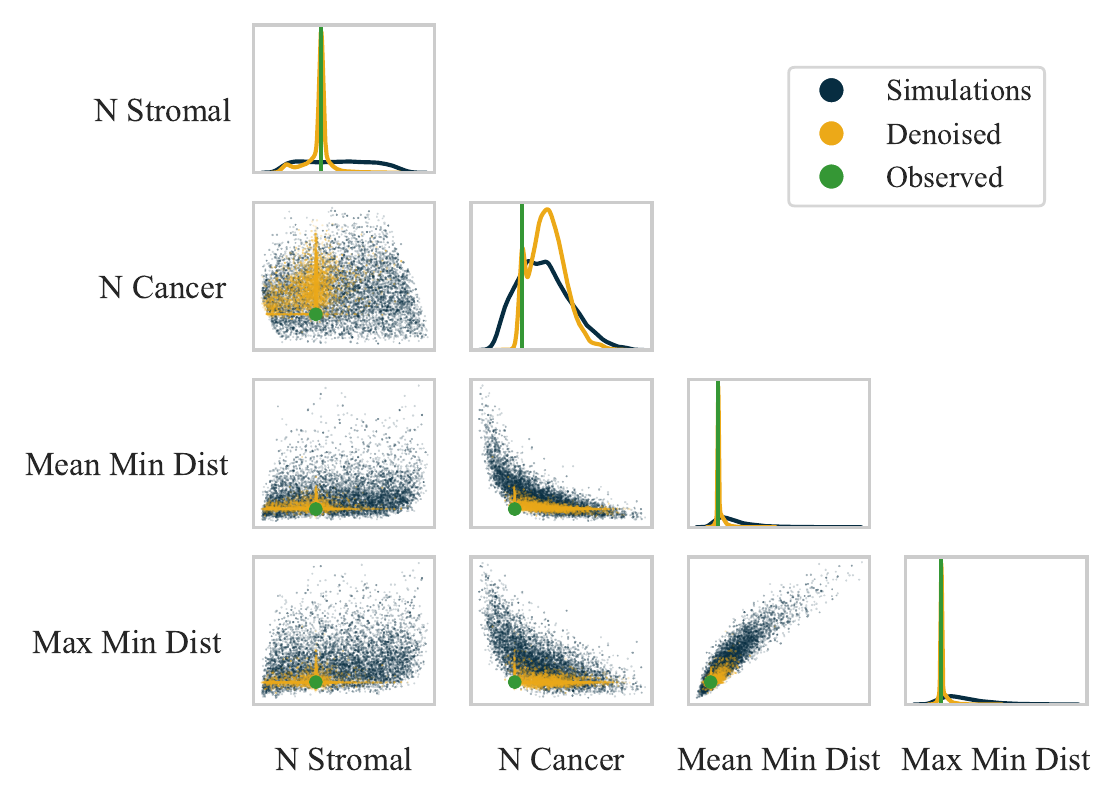}
    \caption{Pair plot of the denoised data for the CS task, corresponding to the example shown in Section \ref{section:results}. See Section \ref{section:tasks} for the summary statistic definitions.}.
    \label{fig:cancer_denoised_pairplot}
\end{figure}

\begin{figure*}[t]
    \centering
\includegraphics[width=0.6\textwidth]{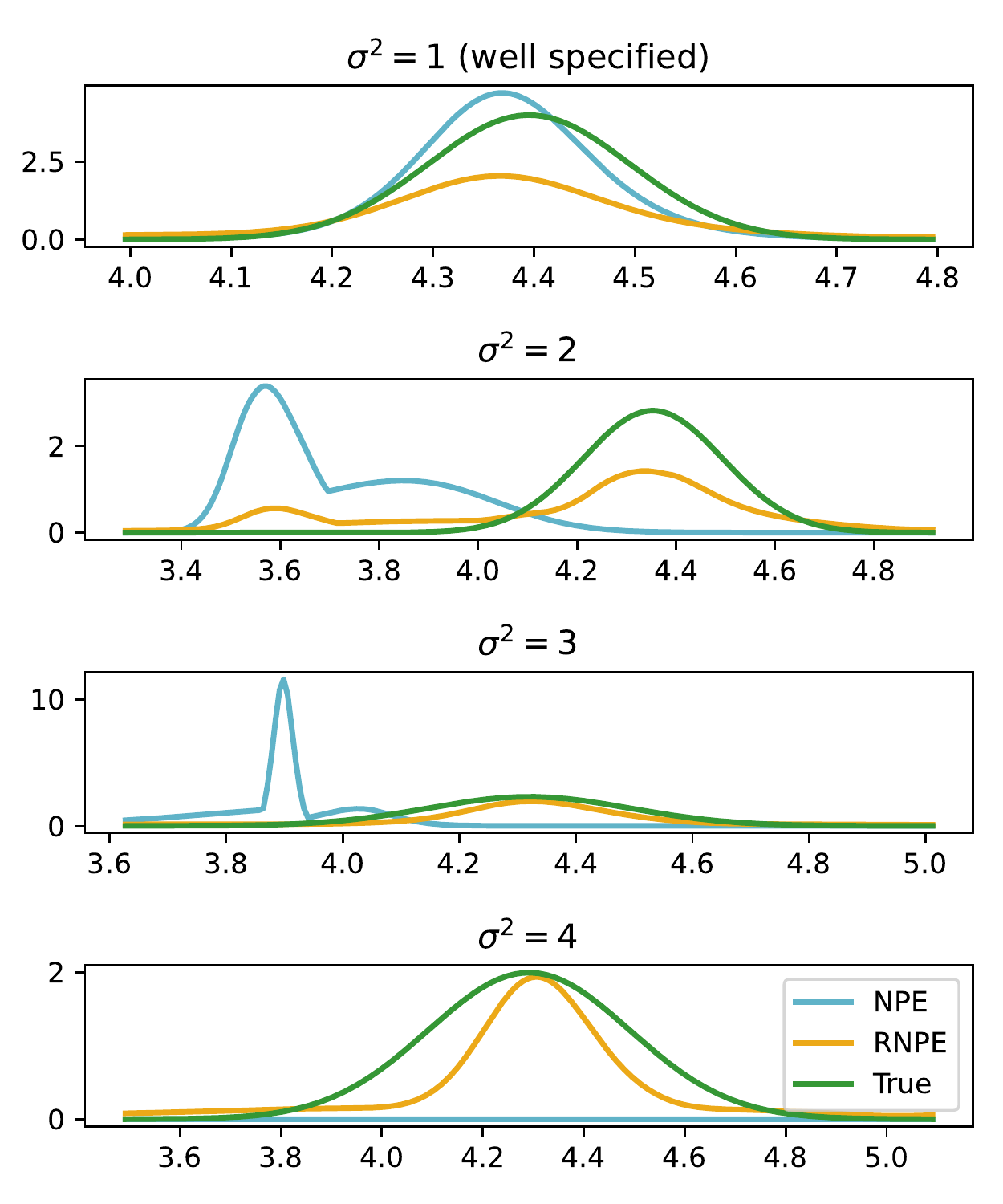}
\caption{\Gls{rnpe} and \gls{npe} under increasing misspecification for the Gaussian task. Note that at $\sigma^2=4$, \gls{npe} failed to put significant mass within $\pm 4$ standard deviations of the true posterior mean. The ``True" posterior is given by the posterior under the true data generating process model.}
\label{fig:gaussian_increasing_misspecification}
\end{figure*}

\end{document}